\documentclass[12pt]{article}

\usepackage[margin=1in]{geometry}
\usepackage{newtxtext,newtxmath}  
\usepackage{setspace}
\usepackage{parskip}              
\usepackage{natbib}

\usepackage{amsmath,amssymb,amsthm}
\usepackage{booktabs,multirow,wrapfig}
\usepackage{algorithm}
\usepackage{algpseudocode}
\usepackage{tikz}
\usepackage{mathtools}
\usepackage{appendix}
\usepackage{url}
\usepackage{titlesec}
\titleformat{\section}
  {\normalfont\large\bfseries}   
  {\thesection}{1em}{}           

\titleformat{\subsection}
  {\normalfont\normalsize\bfseries}
  {\thesubsection}{1em}{}

\titleformat{\subsubsection}
  {\normalfont\small\bfseries}
  {\thesubsubsection}{1em}{}

\titlespacing*{\section}{0pt}{1.5ex plus 1ex minus .2ex}{1ex}
\titlespacing*{\subsection}{0pt}{1.25ex plus 1ex minus .2ex}{0.75ex}
\titlespacing*{\subsubsection}{0pt}{1ex plus .8ex minus .2ex}{0.5ex}

\theoremstyle{plain}
\newtheorem{theorem}{Theorem}
\newtheorem{lemma}{Lemma}
\theoremstyle{definition}
\newtheorem{condition}{Condition}

\title{\vspace{-2em}Bias-Aware Mislabeling Detection via Decoupled Confident Learning}

\author{
  Yunyi Li \\ University of Texas at Austin \\ \texttt{Yunyi.Li@mccombs.utexas.edu}
  \and
  Maria De-Arteaga \\ University of Texas at Austin \\ \texttt{dearteaga@mccombs.utexas.edu}
  \and
  Maytal Saar-Tsechansky \\ University of Texas at Austin \\ \texttt{maytal.saar-tsechansky@mccombs.utexas.edu}
}

\date{}

\begin{document}

\maketitle

\begin{abstract}
Reliable data is a cornerstone of modern organizational systems. A notable data integrity challenge stems from label bias, which refers to systematic errors in a \emph{label}---a covariate that is central to a quantitative analysis---such that its quality differs across social groups. This type of bias has been conceptually and empirically explored, and is widely recognized as a pressing issue across critical domains. However, effective methodologies for addressing it remain scarce. In this work, we propose Decoupled Confident Learning (DeCoLe), a principled machine learning-based framework specifically designed to detect mislabeled instances in datasets affected by label bias, enabling bias-aware mislabelling detection and facilitating data quality improvement. We theoretically justify DeCoLe’s effectiveness and evaluate its performance in the impactful context of hate speech detection, a domain where label bias is a well-documented challenge. Empirical results demonstrate that DeCoLe excels at bias-aware mislabeling detection, consistently outperforming alternative approaches for label error detection. Our work identifies and addresses the challenge of bias-aware mislabeling detection and offers guidance on how DeCoLe can be integrated into organizational data management practices as a powerful tool to enhance data reliability.
\end{abstract}


\section{Introduction}
In an era defined by rapid digitization and fueled by advances in computational technologies and algorithms, organizations are increasingly reliant on data to drive competitiveness, inform strategy, and support core operations. Evidence from industry underscores the transformative nature of data-driven practices, as data-driven organizations outperform peers by a factor of 23 in customer acquisition, achieve ninefold improvements in customer retention, and realize up to 19 times greater profitability \citep{McKinseyDataMatrics, davenport2006competing}. 

Simultaneously, as the strategic value of data-driven practices has intensified, the cost of poor data quality has emerged as a critical point of vulnerability. Analytical systems built on systematic errors in data trigger adverse downstream consequences—producing misleading insights and information that can distort managerial decisions, risk operational disruptions and compliance failures, and erode strategic value and organizational trust \citep{ballou1985modeling, bai2012managing, krishnan2005data, hazen2014data, parssian2004assessing}.  The estimated financial implications of poor data quality are profound, with costs reaching 8\% to 12\% of revenues for typical organizations, potentially leading to billions of dollars in annual losses \citep{hazen2014data}. Data integrity is at the core of contemporary business competitiveness, which increasingly relies on systematic data auditing and relabeling to enhance data quality \citep{ganju2022electronic, liu2024financial, bernhardt2022active}. 

A particularly consequential data integrity challenge stems from the quality of labels \citep{frenay2013classification}. A \emph{Label} is a covariate that is typically the key focus of an analysis, the result of which often informs impactful decisions and policies at scale and advances knowledge discovery. For instance, a medical diagnosis functions as a label that reflects the presence and severity of a condition and is fundamental to inform treatment decisions.  Similarly, in the context of content moderation, a post may be labeled as `toxic' if it contains hate speech or derogatory content~\citep{davani2023hate}, based on which the platform may intervene to mitigate risks. The term ``label" has been coined and used extensively in the machine learning literature, while different terms for this concept---such as dependent variable, response variable, outcome, annotation, or rating---exist across disciplines, including social sciences,  where these terms similarly refer to an outcome of consequence, central to the motivation and implications of the analyses. 

\emph{Label bias} is a particular form of label error that has attracted significant attention across research communities, including Information Systems (IS) and computer science, because it has been shown to be both common and consequential across domains \citep{obermeyer2019dissecting, fu2020artificial, bjarnadottir2020machine, kokkodis2021demand, fu2022fair, li2024label, jacobs2021measurement, passi2019problem, suresh2021framework,sap2019risk, davani2023hate, propublica2016machine,zanger2024risk, hunter1979differential, sjoding2020racial, fogliato2020fairness, akpinar2021effect}.
Label bias refers to systematic errors in the \emph{observed} labels relative to the \emph{gold standard} labels, such that the quality of labels differs across social groups~\citep{li2022more, zanger2024risk}. Such bias can result in disproportionate harms, with some groups being more adversely affected than others. Specifically, under this bias, the likelihood that an instance is incorrectly labeled depends not only on its true label---a dependency that determines what type of error is more common---but also on its group membership (e.g., age bracket, race, gender, or socioeconomic status). A dependency on both true label and group membership means that the types of errors as well as the rates of errors may vary across groups. Label bias commonly arises in settings where the true label is unobservable or costly to obtain, and observed labels are instead derived from proxies, human judgments or measurement instruments that can embed label bias due to social structural biases, cognitive heuristics, or gaps in domain knowledge~\citep{li2022more, li2024label, passi2019problem, jacobs2021measurement,eickhoff2018cognitive, draws2021checklist}.

Prior research has revealed the adverse impact of label bias across systems and analyses that utilize the labels, demonstrating that such errors undermine productive data-driven decisions across domains, and potentially perpetuate existing inequalities \citep{li2024label, li2022more,sap2019risk, davani2023hate}. Analyses that are crucially impacted by labeling quality include data-driven hypothesis generation and pattern discovery, as well as predictive modeling, descriptive modeling, and prescriptive modeling that inform understandings and policies \citep{agarwal2014big, shmueli2010explain, abbasi2024pathways}. These risks have also been echoed by federal agencies and policies, including by the National Institute of Standards and Technology (NIST), which  urged organizations to enhance data quality ~\citep{nist2023ai_rmf}.

It is therefore consequential to advance reliable methods that effectively detect likely mislabeled instances, so that costly auditing, relabeling, and other mitigation efforts can be directed where they are most needed. However, to date, existing methods for mislabeling detection do not consider settings in which mislabeling has the pattern of label bias. This is a problem because label bias is prevalent, especially in high-stakes domains where ground truth is often unavailable and labels are produced via imperfect proxies or based on imperfect human judgments. 

For example, health care systems often aim to estimate health needs to inform the allocation of health resources. However, health needs are often prohibitively costly to estimate. Consequently, health care costs—which are abundant and readily available in the U.S.—have been used as a proxy label for health needs~\citep{obermeyer2019dissecting, li2022more}. However, there is a systematic mismatch between health costs and actual health needs, and this mismatch varies across groups. Specifically, because of historical inequities in access to care across patient groups, Black patients have often incurred lower health care costs than white patients with comparable medical conditions. Consequently, a system that relied on cost as a proxy for needs systematically underestimated Black patients' needs~\citep{obermeyer2019dissecting}. Label quality has been impacted by label bias across a range of other contexts, including criminal justice~\citep{propublica2016machine, zanger2024risk}, content moderation~\cite{davani2023hate, davani2022dealing, sap2019risk}, and hiring~\citep{hunter1979differential}.

In content moderation, research has shown that the accuracy of crowd-sourced hate speech labels differs across groups that are the subjects of hate speech~\citep{davani2023hate}. For example, hate speech targeting the LGBTQ community was found to be at higher risk of being wrongly labeled as non-hateful~\citep{davani2023hate}. 
Such bias in hate speech labels shapes mitigation efforts and thereby undermines the very purpose of safeguarding vulnerable groups.

Despite growing awareness of label bias and its downstream harms, prior work has not proposed scalable and effective methods for identifying mislabeled instances under labeling bias. We refer to this task as \emph{bias-aware mislabeling detection}. Existing studies that address labeling bias have primarily focused on characterizing label bias and its consequences, or on proposing conceptual strategies to avoid it,such as recommending less biased proxy variables~\citep{obermeyer2019dissecting, mullainathan2021inequity, passi2019problem}. However, better, bias-proof proxies are often unavailable, or, if they exist, come at a high cost per instance.
Critically, extensive evidence shows that many widely used and valuable datasets suffer from entrenched label bias ~\citep{li2022more, zanger2024risk, davani2023hate, propublica2016machine, obermeyer2019dissecting, davani2022dealing, sap2019risk}. Effective bias-aware mislabeling detection is thus vital not only for improving the quality of existing datasets but also for guiding label quality assurance in new datasets when bias-free labeling is not feasible to achieve. 

Among methodological work, some prior research has aimed to mitigate particular adverse outcomes of label bias with a focus on supervised machine learning tasks~\citep{jiang2020identifying, wang2021fair, kamiran2009classifying}. However, such methods do not improve the labels' quality, but instead focus on improving learning outcomes in the presence of label bias. Importantly, bias-aware mislabeling detection aims to improve the labeling quality at the outset and  is thus complementary of such methods, as it aims to improve the reliability of labels in the data itself, which can then be used in a variety of downstream applications. This responds to the common organizational reality in which a single dataset is typically used to support a myriad of data-driven analyses and technologies. As such, our work belongs to the literature on mislabeling detection, where we address a critical gap: existing methodologies assume label error structures that are implicitly incompatible with the presence of label bias. 

In this work, we propose and evaluate Decoupled Confident Learning (DeCoLe), a principled machine learning-based framework designed to perform bias-aware mislabeling detection. We first introduce and define the problem of bias-aware mislabeling detection. We then develop DeCoLe, a novel methodology designed to effectively perform bias-aware mislabeling detection. We provide theoretical justification for DeCoLe’s effectiveness in addressing this challenge. We share insights about DeCoLe’s differential performance relative to alternative methods under controlled settings and empirically demonstrate its superior performance in bias-aware mislabeling detection for content moderation—a domain where label bias is a documented challenge. To the best of our knowledge, this is the first empirical evaluation of mislabel detection in the presence of label bias. Crucially, evaluating the effectiveness of mislabeling detection methods is inherently challenging, given that it requires datasets with (1) observed noisy labels, (2) higher-quality gold-standard labels, and (3) demographic group annotations—a combination rarely available in practice due to privacy concerns, organizational liability considerations, and the high cost of collecting high-quality gold-standard labels for large amounts of data. To assess DeCoLe relative to prior mislabeling detection methods, we leverage a rare opportunity afforded by the dataset introduced by \cite{kennedy2020constructing}, which uniquely satisfies these criteria. Empirical results provide robust evidence that DeCoLe substantially outperforms prior methods in bias-aware mislabeling detection for hate speech data. Specifically, DeCoLe identifies more mislabeled instances, especially for the error type that disproportionately affects certain groups, while maintaining high precision of the observed labels in the data estimated as correctly labeled, particularly for the class most impacted by label bias. DeCoLe improves mislabeling detection across all groups without compromising performance for any, challenging the common assumption that correcting bias for some requires sacrificing others and demonstrating that group-aware error detection can benefit all.

The paper proceeds as follows. Section \ref{lit_review} reviews the related literature. Section \ref{methodology} formally defines bias-aware mislabeling detection and introduces the Decoupled Confident Learning (DeCoLe) framework, and offers theoretical support for its performance. Section \ref{experiments} provides empirical insights into DeCoLe's differential effectiveness, and demonstrates its strong bias-aware mislabeling detection capabilities compared to alternative methods in the challenging context of hate speech detection, a domain where label bias remains a persistent problem. We conclude in section \ref{conclusion} with a discussion of DeCoLe's contribution for enhancing data quality, managerial implications, and directions for future research.

\section{Related Literature} \label{lit_review}
In this section, we first review the related work on data integrity and data quality, then discuss label bias, as well as bias mitigation. 

\subsection{Data Integrity and Data Quality}
Ensuring data integrity and data quality control has been a foundational priority in scientific and applied disciplines for decades. Since the 1970s, scholars across fields—including Information Systems(IS), computer science, and data management—have emphasized the significance of trustworthy data as a prerequisite for reliable decision making and organizational performance \citep{adams1975management, ballou1985modeling}. 
These early insights laid the foundation for understanding that high-quality data enables effective decision-making, while poor data quality can undermine outcomes. 

In the era of big data—characterized by the 4Vs (Volume, Velocity, Variety, and Veracity)\footnote{Big data is characterized by the 4Vs: Volume, denoting its massive scale; Velocity, highlighting the rapid generation and need for timely processing; Variety, reflecting diverse data types; and Veracity, emphasizing the importance of data trustworthiness.\citep{lu2020data, cai2015challenges}} and propelled by rapidly advancing information technologies \citep{lu2020data, cai2015challenges}—, the importance of reliable data to unlock the potential of data-driven insights and analytics is paramount~\citep{ McKinseyDataMatrics, davenport2006competing}. Accordingly, there is an expansion in the number of educational programs \citep{jafar2017emergence}, designed to equip graduates with the necessary skills to support organizational decision-making, with a strong emphasis on ensuring data integrity as a fundamental priority ~\citep{lu2020data}. This reflects a growing awareness that maintaining data integrity is not just a best practice, but a necessity. Poor data quality leads to costly inefficiencies, misinformed decisions, and lost opportunities ~\citep{parssian2004assessing,dey2010reassessing,hazen2014data}. 
Ultimately, low-quality data renders information unreliable, undermines the trustworthiness of analytical outputs, compromises usability, and weakens the foundation of decision-support systems \citep{bai2012managing, agarwal2014big, shmueli2010explain}.  Early detection and correction of errors are therefore essential to prevent distortions from compounding throughout the data pipeline. As errors become more entrenched, the cost of correction rises sharply at each subsequent stage of development~ \citep{currim2012modeling}, underscoring the need for organizations to proactively assess and address data quality issues at their root \citep{nist2023ai_rmf}.

Although there is a long-established body of research on data quality, much of it is anchored in the principle of ``fitness for use," emphasizing the role of context in determining data reliability ~\citep{wand1996anchoring}. Data quality is often framed as a multi-dimensional construct, encompassing attributes such as accuracy, completeness, and timeliness. Efforts have been devoted to refining these dimensions and developing methodologies for their measurement~\citep{ballou1985modeling, krishnan2005data, dey2010reassessing}. This stream of work intersects with the ``label construct gap" discussed in the next sub-section, which is one of the sources of label bias. However, it is not always clear how to map the macro view of data concepts to the micro usage of data, where each instance carries its own history and story. Existing research largely lacks a systematic approach for bias-aware mislabeling detection in datasets designed for specific tasks, leaving a crucial gap in ensuring high-quality, unbiased data. To address this gap, we propose a principled machine-learning framework specifically designed to detect mislabeled instances in datasets affected by label bias, offering a robust and systematic solution for facilitating data quality improvement at the instance level.

\subsubsection{Mislabeling detection} 
DeCoLe builds upon a well-established stream of work that focuses on estimating mislabeled instances through principled approaches to noise and error detection~\citep{valiant1984theory, angluin1988learning}. Our work is most closely related to the work by \cite{northcutt2021confident}, which introduces a framework for detecting mislabeled instances via a method we henceforth refer to as Confident Learning (CL)~\citep{northcutt2021confident}. A foundational aspect of this literature is the classification noise process (CNP) introduced by \cite{angluin1988learning}, which assumes that label noise is class-conditional—depending solely on the latent true class. This assumption is reasonable and often appropriate for many datasets ~\citep{goldberger2017training, sukhbaatar2014training}, however, it implicitly rules out the possibility of label bias. In this work, we relax this assumption and propose a methodology for bias-aware mislabeling detection, accounting for the possibility that the likelihood of an erroneous label may depend not only on the latent true class, but also on group membership.

\subsection{Label Bias}
Label bias has gained substantial attention as a key issue affecting data quality. Label bias has been defined as a systematic discrepancy between the outcome of interest and the observed labels, such that the relationship underlying the mismatch varies across groups ~\citep{li2022more, li2024label}. Different sources of label bias have been identified~\citep{li2022more, li2024label}. One significant source of label bias is the \textit{human labeling bias}, which arises when labels reflect annotators’ cognitive biases, prejudices, or limited domain knowledge \citep{howe2008crowdsourcing}. With the rise of crowd-sourcing services~\citep{howe2008crowdsourcing}, researchers have increasingly highlighted the risks posed by annotator cognitive biases~\citep {eickhoff2018cognitive, draws2021checklist} and the potential presence of prejudice in labeling decisions~\citep{otterbacher2015crowdsourcing}. For example, hate speech annotators' judgments are influenced by biased assumptions in language, leading to unfair and disproportionate impacts on marginalized groups ~\citep{davani2023hate}. Expert assessments can also embed biases. In healthcare, for example, it has been shown that the quality of pain assessment and treatment recommendations can be undermined by provider biases ~\citep{hoffman2016racial}. 
Another major source of label bias is \textit{label measurement bias}, which arises when the label being measured aligns conceptually with the construct of interest but is systematically mismeasured across different groups ~\citep{jacobs2021measurement}. This type of bias is prevalent when measurement tools are designed around the needs of one group, overlooking others. A clear example is pulse oximetry, which has been shown to overestimate oxygen saturation levels in patients with darker skin tones. Black patients experience occult hypoxemia—low oxygen levels undetected by pulse oximetry—at nearly three times the rate of white patients \citep{sjoding2020racial}. 

Label bias can also arise from a \textit{label construct gap}, which occurs when there is a discrepancy in theoretical definitions between the construct of interest and the construct observed~\citep{jacobs2021measurement, passi2019problem, li2022more}. This discrepancy often stems from the repurposing of pre-existing datasets, where the original data collection goals differ from the intended task \citep{agarwal2014big}. For example, in healthcare, insurance claim costs often serve as a proxy for patient needs, but due to historical disparities Black patients incur lower expenses despite equal or greater medical needs, leading to inaccurate assessments ~\citep{obermeyer2019dissecting}. Similarly, in college admissions, GPA or class rankings may be used as proxies for student success, ignoring broader factors such as leadership, creativity, and resilience, thereby disadvantaging certain demographics ~\citep{suresh2021framework}. 

Despite its widespread presence, research focused on \emph{label bias} has primarily centered on characterizing or conceptualizing it. \cite{li2022more} empirically demonstrates that collecting more data can exacerbate bias if label bias is overlooked. \cite{fogliato2020fairness} finds that even small biases in observed labels can produce disparate performance across races of recidivism risk assessment tools, and \cite{akpinar2021effect} shows that differential rates in crime reporting can lead to bias in predictive policing systems. DeCoLe advances this literature by moving beyond characterizing label bias to detecting mislabeled instances in datasets potentially affected by it,  providing an algorithmic solution to perform bias-aware mislabeling detection that supports effective mitigation of label bias at scale.

\subsection{Bias Mitigation in Artificial Intelligence (AI)}
The risk that bias embedded in data may yield biased AI systems have gained significant attention in the past years~\citep{barocas2016big,de2022algorithmic}. This has given rise to a set of methodologies to mitigate bias of different forms~\citep{de2022algorithmic}. Existing approaches have sought to address the underrepresentation of certain groups through approaches such as adaptive instance selection~\citep{cai2022adaptive}, synthetic data augmentation~\citep{theodorou2025improving}, or statistical sampling correction~\citep {zelaya2019parametrised}. Another body of work has sought to modify the feature space in order to remove unwanted correlations with protected attributes \citep{zemel2013learning, caliskan2017semantics, bolukbasi2016man, zhao2018learning, bolukbasi2016man, yu2023unlearning, blodgett2020language}. In the context of natural language processing, research has demonstrated that word embedding --- vector representations of words capturing their meanings and relationships--- encodes associations that reflect and reinforce societal biases \citep{caliskan2017semantics, bolukbasi2016man}. To address this, researchers have developed strategies that aim to de-bias embeddings while preserving linguistic integrity \citep{zhao2018learning, bolukbasi2016man, yu2023unlearning, blodgett2020language}. While these efforts address representation issues and feature space bias in data, our work complements this line of research by focusing on label bias.

\subsubsection{Learning from Erroneous and Biased Labels}
A stream of literature on supervised machine learning has aimed to develop methodologies that are robust to errors in the labels used to train them. This includes methods that aggregate labels from multiple annotators using probabilistic models ~\citep{zhang2016learning, snow2008cheap, smyth1994inferring, dawid1979maximum, whitehill2009whose, welinder2010multidimensional, yan2010modeling, zhang2016learning}, approaches that train models directly on noisy labels by adjusting loss functions or model architectures~\citep{shu2019meta, natarajan2013learning, ma2020normalized, lipton2018detecting, van2015learning, patrini2017making, chen2019understanding}, and techniques that manage different misclassification costs via post hoc decision threshold adjustments~\citep{elkan2001foundations}.

The risk of label bias---and in particular the possible presence of group- and class-conditional label errors---has been recognized as a problem by this stream of research, and a few works have proposed methodologies to learn predictive models when such bias is present in the data~\citep{wang2021fair, jiang2020identifying, kamiran2009classifying}.  
These methods have introduced techniques that modify loss functions to enable robust learning in the presence of these types of errors. \cite{wang2021fair} propose fairness-aware surrogate loss functions and constraints. However, the method relies on knowing the group- and class-conditional error rates a priori, which is information that is typically inaccessible in practice. 
\cite{jiang2020identifying} develop an iterative re-weighting method to satisfy fairness constraints, but the selection of instances for up-weighting or down-weighting is based solely on group membership and observed labels, without identifying which instances are actually mislabeled. As a result, the method cannot be used to improve the quality of the data itself. 
\cite{kamiran2009classifying} propose identifying the minimum data perturbation required to enforce or approximate demographic parity, a fairness criterion that assumes equal label distributions across groups. However, this assumption is often violated in real-world contexts where outcome rates differ naturally across populations. In healthcare, for example, certain diseases are more prevalent among specific demographic groups; in social services, some populations are at higher risk of adverse outcomes; and in online content moderation, hate speech may disproportionately target some groups.
Crucially, all these methods pursue specific fairness objectives evaluated at the inference stage and focus on mitigating the consequences of label bias for supervised machine learning tasks. In contrast, DeCoLe addresses label bias directly by performing bias-aware mislabeling detection within the data itself, making it a critical component in facilitating label bias mitigation and data quality improvement. This upstream, data-centric approach not only complements model-level fairness interventions but also improves data quality in a way that supports a broader range of tasks beyond supervised learning.

\section{Methodology} \label{methodology}
In this section, we propose Decoupled Confident Learning (DeCoLe), an algorithm designed to effectively perform bias-aware mislabeling detection. We begin by formally defining the problem of mislabeling detection in Section \ref{preliminary}. Then, in Section \ref{DeCoLe} we present Decoupled Confident Learning (DeCoLe), a principled machine learning framework specifically designed for bias-aware mislabeling detection. Finally, in Section \ref{Theory} we provide a theoretical analysis of DeCoLe's expected performance under both ideal and noisy conditions.

\subsection{Bias-Aware Mislabeling Detection} \label{preliminary}
We consider the problem of detecting mislabeled instances in a dataset that may be affected by label bias, a task we refer to as bias-aware mislabeling detection. Let $\boldsymbol{D} \coloneqq (\boldsymbol{x}, g,\tilde y) ^n$ denote a dataset of $n$ instances, where each instances is represented by features $\boldsymbol{x}$ with associated observed label $\tilde y \in \{0,1\}$ and group membership $g$. Group membership may denote attributes such as age, gender, race, or an intersection of multiple of these. Finally, let $y^*$ represent a latent \emph{gold standard} label. 

We assume there exists a group-and class-conditional noise such that there is a non-correspondence between the observed label $\tilde y$ and the latent gold standard label $y^*$
that depends on both the gold standard label, $y^*$, and the group membership, $g$. 
Let $\pi_{c, g_k}$ denote the misclassification rate for observed labels $\tilde{y}=c$. Specifically, for a particular group $g_k$, where $k \in \{0,..., (K-1)\} $ and $K$ denotes the total number of possible groups, positive instances may be misclassified as negative with a probability $\pi_{0,{g_k} } = p(\tilde y = 0| y^* = 1, g =g_k)$, and negative instances may be misclassified as positive with a probability $ \pi_{1,{g_k} } = p(\tilde y = 1| y^* = 0, g =g_k)$. Label bias occurs when there is a disparity in one (or both) of these metrics across different groups $g_k$ ~\citep{li2022more}. We formulate the problem under a binary labeling setting, while allowing for multi-categorical group memberships. Our goal is to provide a generic framework that can be used to identify mislabeled instances in a dataset that is potentially affected by label bias.


\begin{table}[ht]
    \centering
    \caption{Key Notations in DeCoLe}
    \label{tab:descriptive-stats}
    \begin{tabular}{l|l}
        \hline
        \hline
        Notation & Description  \\
        \hline
        $ \tilde y$ & Observed (possibly erroneous) labels; binary label. \\
         $y^*$ & Gold standard labels; binary label. \\
         $g$ & Group indicator; a categorical attribute, $g_k \in \{0,...,(K-1)\}$ denotes a distinct group.\\
        $\boldsymbol{D}_{\tilde y = i, y^* = j}$ & The set of instances such that $ \tilde y=i$ and $y^*=j$.   \\  
        $\boldsymbol{D}_{g_k}$ & The set of instances belonging to group $g_k$.\\
         $\pi_{c, g_k}$ & The misclassification rate for observed labels $\tilde{y} = c$, for $c \in \{0,1\}$, and group $g=g_k$.\\
         $\boldsymbol{X}_{g_k} $ &  The set of feature vectors for all instances in group $g_k$.\\ 
        $\hat p(\boldsymbol{x}_{g_k})$ & The probability predicted by classifier $f_k$ for $\boldsymbol{x}_{g_k} \in \boldsymbol{X}_{g_k}$ to be in the positive class. \\
    CNS$_{g_k}$ & Confident Negative Set for group $g_k$ $\forall \boldsymbol{x}_{g_k} \in $ CNS$_{g_k}$, $\hat{ y^* } = 0 $, defined in Eq.\ref{eq:cns}.  \\
        CPS$_{g_k}$ & Confident Positive Set for group $g_k$ $\forall \boldsymbol{x}_{g_k} \in $ CPS$_{g_k}$, $\hat{ y^* } = 1 $, defined in Eq.\ref{eq:cps}.  \\
        UB$_{g_k}$ & Upper Bound threshold for constructing CNS$_{g_k}$, defined in Eq.\ref{eq:ub}.\\
        LB$_{g_k}$ & Lower Bound threshold for constructing CPS$_{g_k}$, defined in Eq.\ref{eq:lb}.  \\
        $\hat{y^*} $ & Confident predictions; for instances belong to confident sets only.\\
        $\boldsymbol{D}_{\tilde y \neq y^*|g}$ & Mislabeled instances.\\
        $\boldsymbol{\hat{D}}_{\tilde y \neq y^*|g}$ & Instances estimated to be mislabeled by DeCoLe applied considering group $g$.\\
        $\boldsymbol{D}_{\tilde y = y^*|g}$ & Correctly labeled instances. \\
        $\boldsymbol{\hat{D}}_{\tilde y = y^*|g}$ & Instances estimated to be correctly labeled by DeCoLe applied considering group $g$. \\
        \hline
    \end{tabular}
\end{table}

\subsection{Decoupled Confident Learning (DeCoLe)} \label{DeCoLe}
\begin{figure}[h]
\centering\includegraphics[scale = 1.25]{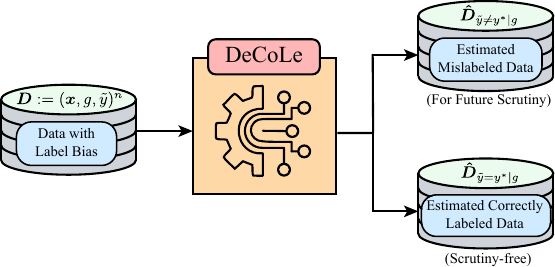} 
\caption{Overview of DeCoLe. Given data with label bias, DeCoLe identifies a subset of confidently labeled data to be used without further scrutiny, while flagging potentially mislabeled instances for targeted review—thereby enabling bias-aware data quality improvement.}
\label{fig:DeCoLe_abstract}
\end{figure}

We propose Decoupled Confident Learning (DeCoLe) to effectively perform bias-aware mislabeling detection. DeCoLe only needs access to the observed dataset $\boldsymbol{D} \coloneqq (\boldsymbol{x}, g, \tilde y) ^n$, and does not require access to any gold standard labels $y^*$. The input to DeCoLe is the original data, $\boldsymbol{D}$, and the output is a partition of the data into two sets, one estimated to be mislabeled, and the other one estimated to be correctly labeled, as shown in Figure~\ref{fig:DeCoLe_abstract}. 

At a high level, DeCoLe addresses the challenge of bias-aware mislabeling detection by separately performing a series of confident learning procedures for each group. This group-specific approach enables independent estimation of the joint distribution between observed biased labels and latent gold standard labels, effectively detecting label errors under group- and class-conditional noise. In doing this, DeCoLe builds on the common machine learning principle that predicted probabilities reflect confidence in classifications. This idea is formalized in the concept of  ``probably approximately correct identification'' ~\citep{valiant1984theory}, which, within the theory of learnability, states that after randomly observing both positive and negative instances, an identification procedure should infer a hypothesis that, with high probability, is close to the correct concept~\citep{angluin1988learning, valiant1984theory}. In our approach, we incorporate the insight that the underlying concept may differ across groups ~\citep{dwork2018decoupled, gupta2023same}. As a result, effective identification requires inferring group-specific hypotheses, each guided by its own tailored identification procedure.

Specifically, for each group $g_k$, DeCoLe first generates out-of-sample predicted probabilities $\hat p(\boldsymbol{x}_{g_k})\: \forall \: x_{g_k} \in \boldsymbol{D}$ —the estimated likelihood that an instance belongs to class 1—produced by a separate classifier $f_k$ trained for group $g_k$. Then, DeCoLe uses the predicted probabilities to compute two \emph{thresholds}: the Lower Bounds (LB$_{g_k}$) of the confident positive set CPS$_{g_k}$, and Upper Bounds (UB$_{g_k}$) of the confident negative set CNS$_{g_k}$. Intuitively, the algorithm is fairly confident in its predictions above and below these bounds, respectively. We then assign a confident prediction $\hat{ y^* } = 1 $ to all instances in the CPS$_{g_k}$, and assign $\hat{y^*} = 0 $ to all instances in the CNS$_{g_k}$.An instance is flagged for future scrutiny if $\tilde{y}\neq \hat{y}^*$. 

The main identification procedure consists of three steps: 
\begin{enumerate}
    \item Train a separate predictive model $f_k$ for each group $g_k$, and obtain out-of-sample predicted probabilities $\hat p(\boldsymbol{x}_{g_k}) = \hat p(\tilde y = 1; \boldsymbol{x}_{g_k}, f_k) \: \forall \: x_{g_k} \in \boldsymbol{D}$ , where $\boldsymbol{x}_{g_k}$ denotes $\boldsymbol{x}$ for $(\boldsymbol{x}, \tilde y) \in \boldsymbol{D}_{g_k}$.
     This process yields predicted probabilities $\hat p(\boldsymbol{x})$ for every instance in the dataset.
    \item For each group $g_k$, estimate upper bound (UB$_{g_k}$)  threshold for Confident Negative Set (CNS$_{g_k}$) and lower bound (LB$_{g_k}$) threshold for Confident Positive Set (CPS$_{g_k}$):
\begin{align}
\text{LB}_{g_k} = \frac{1}{|D_{\tilde y = 1, g = g_k}|}\sum_{\boldsymbol{x}_{g_k}\in \boldsymbol{D}_{\tilde y = 1}} \hat p(\boldsymbol{x}_{g_k}),\label{eq:lb}\\
\text{UB}_{g_k} = \frac{1}{|D_{\tilde y = 0, g = g_k}|}\sum_{\boldsymbol{x}_{g_k}\in \boldsymbol{D}_{\tilde y = 0}} \hat p(\boldsymbol{x}_{g_k}). \label{eq:ub} 
\end{align}
Then, use these bounds to generate the confident sets (CPS$_{g_k}$ and CNS$_{g_k}$) that can be used to identify instances that are inferred to have an erroneous observed label. Formally,
\begin{align}
\text{CPS}_{g_k} = \left\{ (\boldsymbol{x}, \tilde y) \in \boldsymbol{D}_{g_k} : \hat p(\boldsymbol{x}) \geq  \text{LB}_{g_k} \right\}\label{eq:cps} \\
\text{CNS}_{g_k} = \left\{ (\boldsymbol{x}, \tilde y) \in \boldsymbol{D}_{g_k}: \hat p(\boldsymbol{x}) \leq  \text{UB}_{g_k} \right\}.\label{eq:cns} 
\end{align}

\item Instances $(\boldsymbol{x}, \tilde y) \in \boldsymbol{D}_{g_k}$ with observed label $\tilde y = 0$ that fall within the confident positive set, CPS$_{g_k}$, are included in set $\boldsymbol{\hat{D}}_{\tilde y \neq y^*|g}$. Similarly, instances $(\boldsymbol{x}, \tilde y) \in \boldsymbol{D}_{g_k}$ with observed label $\tilde y = 1$ that fall within the confident negative set, CNS$_{g_k} $, are also added to  $\boldsymbol{\hat{D}}_{\tilde y \neq y^*|g}$.
\end{enumerate}


\begin{algorithm}[H]
\caption{Decoupled Confident Learning (DeCoLe)}\label{alg}
    \begin{algorithmic}
    \State {\bfseries Input:} A dataset with noisy labels $\boldsymbol{D} \coloneqq (\boldsymbol{x}, g, \tilde y) ^n$, initialize a set of classifier $\{f_1,..., f_k\}$. 
    \For{$k=0$ {\bfseries to} $(K-1)$}
    \State{Initialize a group-specific bias detection set $\boldsymbol{\hat{D}}_{\tilde y \neq y^*|g = g_k} =  \emptyset $}
    \State {\bfseries Part 1: Estimating $p(\boldsymbol{x})$.}
    \State $f_k$.fit$(\boldsymbol{D}_{g_k})$ 
    \State $\hat p(\boldsymbol{X}_{g_k}) \leftarrow $ $f_k$.predict\_crossval\_prob $(\tilde{y} = 1| \boldsymbol{X}_{g_k})$ where $ \boldsymbol{X}_{g_k} : =  \left\{ \boldsymbol{x} : (\boldsymbol{x}, \tilde{y}) \in \boldsymbol{D}_{g_k} \right\}$
    \State {\bfseries Part 2: Estimating CPS$_{g_k}$ and CNS$_{g_k}$.}
    \State CPS$_{g_k} = \left\{ (\boldsymbol{x}, \tilde y) \in \boldsymbol{D}_{g_k} : \hat p(\boldsymbol{x}) \geq  \text{LB}_{g_k} \right\}$; with LB$_{g_k} = \frac{1}{|D_{\tilde y = 1, g = g_k}|}\sum_{\boldsymbol{x}_{g_k}\in \boldsymbol{D}_{\tilde y = 1}} \hat p(\boldsymbol{x}_{g_k})$
    \State CNS$_{g_k} = \left\{ (\boldsymbol{x}, \tilde y) \in \boldsymbol{D}_{g_k}: \hat p(\boldsymbol{x}) \leq  \text{UB}^*_{g_k} \right\}$; with UB$_{g_k} = \frac{1}{|D_{\tilde y = 0, g = g_k}|}\sum_{\boldsymbol{x}_{g_k}\in \boldsymbol{D}_{\tilde y = 0}} \hat p(\boldsymbol{x}_{g_k})$
    \State {\bfseries Part 3: Identify erroneous instances for future scrutiny.}
    \State $\boldsymbol{ \hat{D}}_{\tilde y \neq y^*|g = g_k} \coloneqq  \boldsymbol{\hat{D}}_{\tilde y \neq y^*|g = g_k} \cup \{( \boldsymbol{x}_{g_k},\tilde{y}) \in $ CPS$_{g_k} $: $\tilde y=0 \}$
    \State $\boldsymbol{ \hat{D}}_{\tilde y \neq y^*|g= g_k} \coloneqq  \boldsymbol{\hat{D}}_{\tilde y \neq y^*|g = g_k} \cup \{(\boldsymbol{x}_{g_k},\tilde{y}) \in $ CNS$_{g_k}$: $\tilde y=1\}$
    \EndFor
    \State $\boldsymbol{\hat{D}}_{\tilde y \neq y^*|g } \coloneqq \{\boldsymbol{ \hat{D}}_{\tilde y \neq y^*|g= g_k}\} ^K$
    \State \Return $ \boldsymbol{\hat{D}}_{\tilde y \neq y^*|g}$
    \end{algorithmic}
\end{algorithm}

The detailed algorithm can be found in Algorithm~\ref{alg}. When there is group and class-conditional noise, decoupling---training individual models for each group---disentangles the group-specific noise structure, and accounts for differential subgroup validity~\citep{gupta2023same}. By training separate classifiers for each group, DeCoLe allows the likelihood of an instance belonging to a specific class to be estimated in a way that reflects the unique characteristics and optimization needs of the group, reducing the tension between optimizing for one group at the expense of another. Importantly, such an approach allows us to accommodate different error patterns across groups: for example, one group may be more susceptible to incorrect observed labels in the form of false positives, while another group is more susceptible to incorrect observed labels in the form of false negatives. 

\begin{figure}[h]
\centering\includegraphics[scale = 0.225]{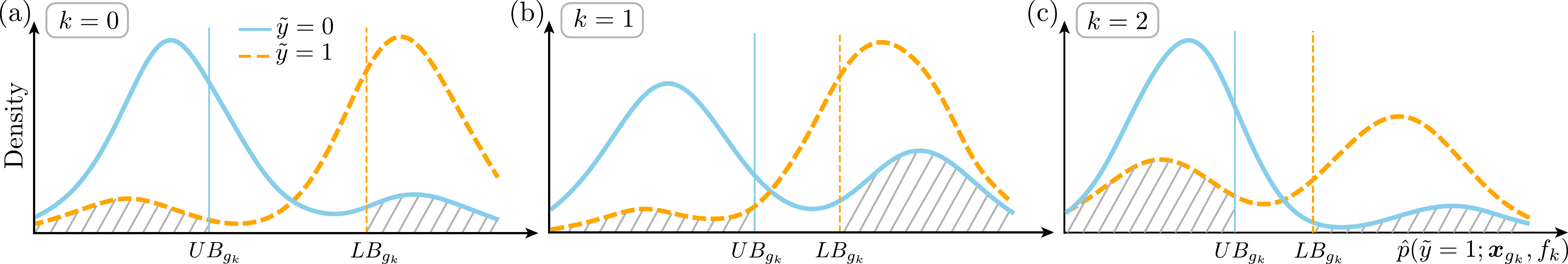} 
\caption{Illustration of DeCoLe's error identification process. Each panel shows predicted probabilities under group-specific models $f_k$, with density curves for observed labels $\tilde{y}=1$ (dashed orange) and $\tilde{y}=0$ (solid sky blue). Shaded regions indicate instances flagged for scrutiny using group-specific thresholds (UB${g_k}$ and LB${g_k}$). Label error patterns differ by group: Group 0 has balanced errors, Group 1 is dominated by false negatives, and Group 2 by false positives—demonstrating group- and class-conditional noise.}
\label{fig:method_fig}
\end{figure}

Figure~\ref{fig:method_fig} illustrates how DeCoLe performs bias-aware mislabeling detection in a binary classification setting. Mislabeled instances are identified using group-specific thresholds tailored to each subgroup. Each panel in the figure corresponds to a different group $g_k$, revealing how error patterns vary across groups —for example, Group 0 exhibits balanced errors, while Groups 1 and 2 are dominated by false negatives and false positives, respectively. This variation highlights the group-and class-conditional nature of label bias, which DeCoLe addresses through adaptive, group-specific error detection.

In the next section, we examine sufficient conditions where DeCoLe perfectly detects all mislabeled instances, even when $\hat p(\boldsymbol{x})$ is noisy.

\subsection{Theoretical Analysis}\label{Theory}
In this section, we examine the sufficient conditions under which $\boldsymbol{\hat {D}}_{\tilde y \neq y^*|g}$ is a consistent estimator for $\boldsymbol{D}_{\tilde y \neq y^*|g}$. We show that, 1) when there is no model noise, and the model $f_{i}: \boldsymbol{x}_{g_k} \rightarrow \hat p(\boldsymbol{x}_{g_k})$ produces ideal predicted probabilities $\hat p(\boldsymbol{x}_{g_k}) $ (defined in Condition \ref{condition1}),  or 2) when the model is imperfect and produces per-instance diffracted noisy predicted probabilities (defined in Condition \ref{condition2}), DeCoLe exactly identifies mislabeled instances in a dataset with label bias. To disambiguate label noise and model noise, we consider two levels of assumption on model noise and the predicted probability $\hat p(\boldsymbol{x}_{g_k}) $, and we show that under both ideal and noisy $\hat p(\boldsymbol{x}_{g_k})$, DeCoLe captures erroneous instances in a dataset with label bias and consistently estimates the set of mislabeled instances $\{ \boldsymbol{D}_{\tilde y \neq y^*|g}\}$.  We demonstrate this in the series of theorems below. We also note that we complement these theoretical guarantees with empirical results in Section~\ref{experiments}, where we show that the proposed method yields state-of-the-art results even when the theoretical conditions identified in this section are violated. 

We begin by considering an ideal scenario that builds on a well-established assumption in the confident learning literature, where predicted probabilities are modeled to capture the class-conditional noise structure for detecting label noise~\citep{northcutt2021confident}. We extend this framework to account for group- and class-conditional noise, acknowledging that label noise can arise not only from class-dependent errors but also from systematic biases across groups. Building on the confident learning literature and under our extended framework, Condition 1 assumes that the predicted probabilities $\hat p(\boldsymbol{x}_{g_k}) $ accurately reflect the underlying group- and class-conditional noise in the labels. We show that, under Condition 1, DeCoLe exactly detects mislabeled instances in a dataset with group and class conditional noise (label bias) and that $\boldsymbol{\hat {D}}_{\tilde y \neq y^*|g}$ is a consistent estimator for $\boldsymbol{D}_{\tilde y \neq y^*|g}$.

Let $\hat p(\tilde y = 1; \boldsymbol{x}_{g_k}, f_k)$ denote the probability predicted by classifier $f_k$ that an instance $\boldsymbol{x}_{g_k} $ belongs to a positive class. And let $\hat p(\boldsymbol{x}_{g_k})$ be the shorthand writing for $\hat p (\tilde y = 1; \boldsymbol{x}_{g_k}, f_k)$.
\begin{condition}[Ideal Predicted Probability]\label{condition1}
When the predicted probability $\hat p(\boldsymbol{x}_{g_k})$ is ideal, we have $\hat p(\boldsymbol{x}_{g_k}) = \hat p (\tilde y = 1; \boldsymbol{x}_{g_k}, f_k) = \hat p (\tilde y = 1; \boldsymbol{x}_{g_k} \in \boldsymbol{D}_{y^* = r, g = g_k}, f_k) =  p (\tilde y = 1; \boldsymbol{x}_{g_k} \in \boldsymbol{D}_{y^* = r, g = g_k}, f_k)  =p (\tilde y  = 1| y^*  = r, g = g_k) = p(\boldsymbol{x}_{g_k})$. Essentially, ideal predicted probability $\hat p(\boldsymbol{x}_{g_k})$ match the noise rates $p(\tilde y  = 1| y^*  = r, g = g_k)$ corresponding to true label $r$ for $\boldsymbol{x}_{g_k}$.
\end{condition}

\begin{theorem} \label{thm1}
For a dataset $\boldsymbol{D} \coloneqq (\boldsymbol{x}, \tilde y) ^n$ with group- and class-conditional label noise $\pi_{0,{g_k}} = p(\tilde y = 0| y^* = 1, g = g_k) < 0.5 $ and $\pi_{1,{g_k}} = p(\tilde y = 1| y^* = 0, g = g_k) <0.5 $, if the classifiers $f_k$ produce ideal predicted probabilities, then the set of detected mislabeled instances $\boldsymbol{\hat {D}}_{\tilde y \neq y^*|g}$ is a consistent estimator for $\boldsymbol{D}_{\tilde y \neq y^*|g}$.
\end{theorem}

While Condition \ref{condition1} provides a useful sanity check, we now relax this assumption to capture a more nuanced and realistic scenario where the noisy predicted probability $\hat p(\boldsymbol{x}_{g_k})$  can vary at the instance level. This assumption is well-established and widely adopted in prior studies (e.g. \cite{jiang2020beyond, northcutt2021confident}). We extend this perspective to explicitly incorporate both group- and class-conditional noise structures. Under this extended framework, we show that DeCoLe successfully identifies mislabeled instances in a dataset with label bias and that $\boldsymbol{\hat {D}}_{\tilde y \neq y^*|g}$ remains a consistent estimator for $\boldsymbol{D}_{\tilde y \neq y^*|g}$.

\begin{condition}[Per-instance diffracted (Noisy) Predicted Probability]\label{condition2}
The predicted probability $\hat p(\boldsymbol{x}_{g_k})$ provided by model $f_k$ is per-instance diffracted if it follows the relationship $\hat p(\boldsymbol{x}_{g_k}) = p(\boldsymbol{x}_{g_k}) + \epsilon_{\boldsymbol{x}_{g_k}} $ where the noise term  $\epsilon_{\boldsymbol{x}_{g_k}} $ is drawn from the following distribution: $\epsilon_{\boldsymbol{x}_{g_k}} \sim \mathcal{U}[\epsilon_k +\text{LB}^*_{g_k}-p(\boldsymbol{x}_{g_k}), \epsilon_k-\text{LB}^*_{g_k}+ p(\boldsymbol{x}_{g_k})] \hspace{0.2em}$   when $ p(\boldsymbol{x}_{g_k})>\frac{1}{2} $; and  $\epsilon_{\boldsymbol{x}_{g_k}} \sim \mathcal{U}[\epsilon_k -\text{UB}^*_{g_k}+p(\boldsymbol{x}_{g_k}), \epsilon_k+\text{UB}^*_{g_k}-p(\boldsymbol{x}_{g_k})]$ when $p(\boldsymbol{x}_{g_k})<\frac{1}{2} $. Here, $\mathcal{U}$ denotes a uniform distribution, and $\epsilon_k = \mathbb{E}_{\boldsymbol{x_{g_k}}}[\epsilon_{\boldsymbol{x}_{g_k}}] $, where $\epsilon_{\boldsymbol{x}_{g_k}}$ represents the deviation from the ideal predicted probability. $\text{LB}^*_{g_k}$ and $\text{UB}^*_{g_k}$ denote the value of $\text{LB}_{g_k}$ and $\text{UB}_{g_k}$ under condition 1.
\end{condition}

\begin{theorem} \label{thm2}
For a dataset $\boldsymbol{D} \coloneqq (\boldsymbol{x}, \tilde y) ^n$, with group- and class-conditional label noise $\pi_{0,{g_k}} = p(\tilde y = 0| y^* = 1, g = g_k) < 0.5 $ and $\pi_{1,{g_k}} = p(\tilde y = 1| y^* = 0, g = g_k) <0.5 $, if the classifiers $f_k: \boldsymbol{x}_{g_k} \rightarrow \hat p(\boldsymbol{x}_{g_k})$ yields per-instance diffracted predicted probabilities, then the set of detected mislabeled instances $\boldsymbol{\hat {D}}_{\tilde y \neq y^*|g}$ is a consistent estimator for $\boldsymbol{D}_{\tilde y \neq y^*|g}$.
\end{theorem}
The theoretical results establish that DeCoLe reliably performs bias-aware mislabeling detection under well-defined conditions. First, when the model produces ideal predicted probabilities, DeCoLe perfectly captures all erroneous labels. Then, relaxing this assumption, we consider the more realistic case where the predicted probabilities are per-instance diffracted (noisy), demonstrating that DeCoLe consistently identifies mislabeled instances. The detailed proofs of these theorems are provided in Appendix \ref{Appdix:Proof}.

\section{Experiments}\label{experiments}
To empirically evaluate the effectiveness of DeCoLe, we first examine its performance in a synthetic setting, where the relationship between the latent construct of interest ($y^*$) and observed labels ($\tilde{y}$) is fully controlled. This illustrates DeCoLe's contribution to effective bias-aware mislabeling detection, and demonstrates how it differs from alternative methods. Building on this foundation, we evaluate DeCoLe and competing algorithms on a real-world hate speech detection context, where the presence and impact of label bias have been widely acknowledged in prior research.

\subsection{Evaluation Metrics} \label{Evaluation}
As illustrated in Figure~\ref{fig:DeCoLe_abstract}, DeCoLe partitions the dataset into two disjoint subsets: (1) estimated mislabeled data, denoted by $\boldsymbol{\hat{D}}{\tilde{y} \neq y^*|g}$ and (2) estimated correctly labeled data, denoted by $\boldsymbol{\hat{D}}{\tilde{y} = y^*|g}$. This decomposition forms the foundation for evaluating DeCoLe’s performance. To quantify performance, we define four evaluation metrics:

\paragraph{1. Overall Recall of Mislabeled Instances.} First, for data estimated to be mislabeled, we assess the proportion of truly mislabeled instances (denoted by $\boldsymbol{D}_{\tilde y \neq y^*|g}$) that are successfully identified for scrutiny. This corresponds to the overall recall of mislabeled instances, and is defined in Equation~\ref{Recall}.
\begin{align}\label{Recall}
\textit{Recall}_{\boldsymbol{\hat {D}}_{\tilde y \neq y^*|g}}
= \frac{|\boldsymbol{D}_{\tilde y \neq y^*|g} \cap \boldsymbol{\hat{D}}_{\tilde y \neq   y^* |g}| }{|\boldsymbol{D}_{\tilde y \neq y^*|g}|}
\end{align}

\paragraph{2. Overall Precision in Data Estimated as Correctly Labeled.} Second, for the data estimated to be correctly labeled, we assess the fraction of these are indeed correctly labeled. This corresponds to the overall precision of the observed labels in the subset estimated to be correctly labeled.  This is defined in Equation~\ref{Precision}.
\begin{align}\label{Precision}
\textit{Precision}_{\boldsymbol{\hat{D}}_{\tilde y = y^*|g}}  = \frac{|\boldsymbol{D}_{\tilde y = y^*|g } \cap \boldsymbol{\hat{D}}_{\tilde y = y^*|g}|}{|\boldsymbol{\hat{D}}_{\tilde y = y^*|g}|}
\end{align}

\paragraph{3. Recall of Bias-Inducing Error.} 
Moreover, because our motivation is grounded on the risks of label bias in a given context, we pay special attention to the error type that induces such bias. We refer to this as \textit{bias-inducing error} — a form of labeling error that is unevenly distributed across groups and has consequential implications within a given context.  
For example, consider a dataset exhibiting label bias, where one group experiences a disproportionate number of false positives. In such cases, we evaluate DeCoLe's effectiveness to in detecting those specific erroneous labels that contribute to group-dependent harm. Specifically, we measure the recall of bias-inducing mislabels. Formally, for a label value $c \in \{0,1\}$, let $\hat {D}_{\tilde y = c,  y^* = \overline{c}|g}$  be the estimated set of instances that is mislabeled as $c$, for which the correct label is $\overline{c}$. The \textit{recall of bias-inducing error} is defined in Equation \ref{Recall_bias}. 
\begin{align}\label{Recall_bias}
\textit{Recall}_{\boldsymbol{\hat {D}}_{\tilde y = c,  y^* = \overline{c}|g}}
= \frac{|\boldsymbol{D}_{\tilde y = c,  y^* = \overline{c}|g} \cap \boldsymbol{\hat{D}}_{\tilde y = c,  y^* = \overline{c}|g}| }{|\boldsymbol{D}_{\tilde y = c,  y^* = \overline{c}|g}|}
\end{align}

\paragraph{4. Precision of Bias-Dominant Class in Data Estimated as Correctly Labeled.} Additionally, we assess the \textit{precision of the bias-dominant class} within the subset of data estimated to be correctly labeled. This corresponds to the class that is affected by bias-inducing errors. For example, when the false positive is the primary form of bias, we evaluate the precision of the positive class in the subset estimated to be correctly labeled. This metric helps us to assess the degree to which group-conditional bias remains in the data that is deemed to be accurate. 
Formally, the \textit{precision of bias-dominant class} for class $c$ is defined in Equation \ref{Precision_bias}.
\begin{align}\label{Precision_bias}
\textit{Precision}_{\boldsymbol{\hat {D}}_{\tilde y = y^* = c|g}}
= \frac{|\boldsymbol{D}_{\tilde y = y^* = c|g} \cap \boldsymbol{\hat{D}}_{\tilde y = y^* = c|g}| }{|\boldsymbol{D}_{\tilde y = y^* = c|g}|}
\end{align}

We refer to Metrics 1 and 2 as measures of \textbf{overall performance}, and Metrics 3 and 4 as measures of \textbf{bias-aware} performance in the context of bias-aware mislabeling detection. We use these two terms throughout Section \ref{experiments} to frame and discuss the results.

\subsection{Evaluation Under Controlled Settings} \label{sec:synthetic}
In this section, we explore and analyze results on a synthetic dataset, where complete control over the relationship between $y^*$ and $\tilde{y}$ allows us to gain insight into DeCoLe's effectiveness in bias-aware mislabeling detection and its performance gains over alternatives.

\subsubsection{Data Generation}\label{Simu_describ}
We create a synthetic dataset with group and class conditional noise rates, enabling full control over the relationship between observed labels $\tilde y$ and latent gold standard labels $y^*$. The synthetic population consists of $N = 10{,}000$ instances, each associated with covariates $\boldsymbol{X} \in \mathbb{R}^2$, a binary group membership $g \in \{0, 1\}$, a latent good standard label $y^* \in \{0, 1\}$, and an observed label $\tilde y \in \{0, 1\}$. We consider group imbalance, a frequent occurrence in real-world data~\citep{mitchell2018prediction}, by assigning  70\% of the population to a predominant group ($g = 1$). Additionally, we account for differential subgroup validity~\citep{hunter1979differential, de2022algorithmic}, a common phenomenon in which the relationship between covariates and target labels varies across groups. To do this, we draw covariates for different group and class combinations from bi-dimensional normal distributions with different means. We use the same standard deviation for all normal distributions. Further details about the sampling of $\boldsymbol{X} \in \mathbb{R}^2$ and $y^* \in \{0, 1\}$ are provided in Appendix \ref{Appdix:Data_Generation}.

We generate observed labels $\tilde y$ with group and class-conditional noise, i.e. different error types for different groups. Suppose the positive class represents opportunities or goods, such as job offers. We assume group $g_0$, the minority group, is more likely to be affected by false negative labels, and group $g_1$, the majority group, benefits from false positive labels. We set $\pi_{1,{g_0}} = 0.4$ and $\pi_{0,{g_1}} = 0.2$, considering group population differences. Additionally, we assume some level of noise for the remaining instances, and set $\pi_{0,{g_0}} = 0.05$ and $\pi_{1,{g_1}} = 0.05$.

\begin{figure}[h] 
    \centering
    \includegraphics[width=0.45\columnwidth]{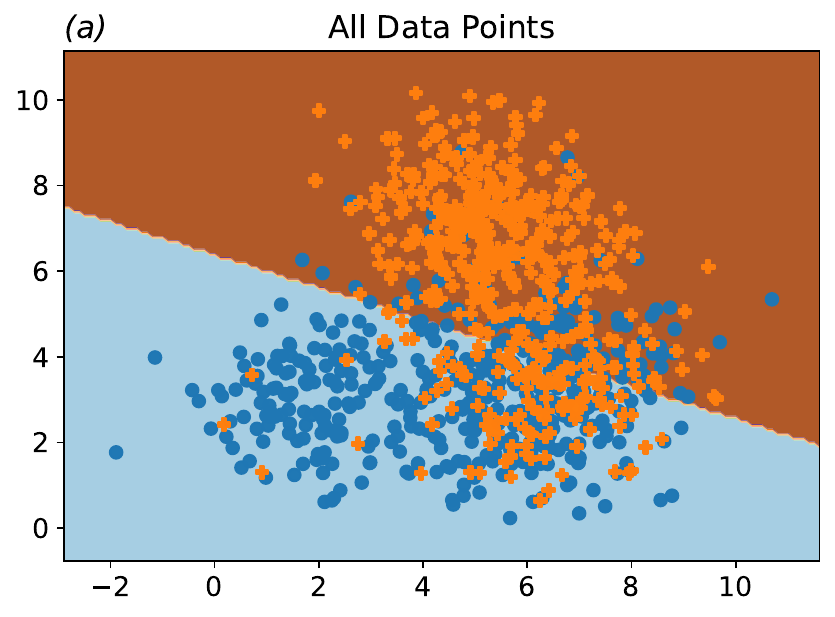}
    \begin{minipage}[c]{0.52\columnwidth}
     \raggedleft
    \includegraphics[scale = 0.36]{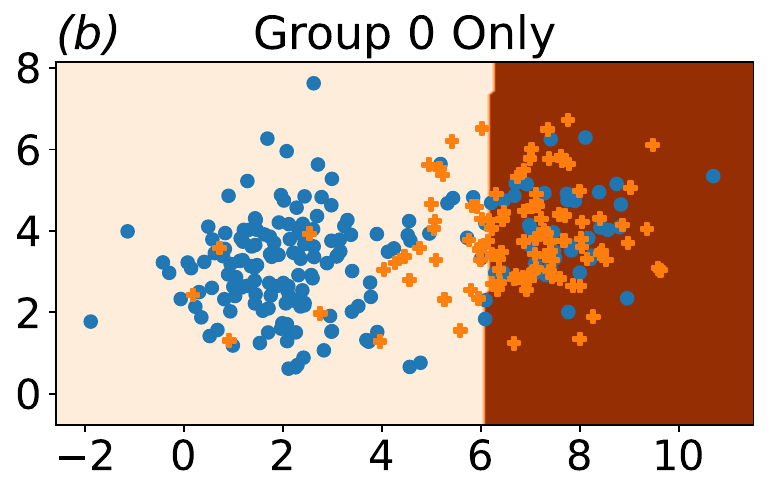} 
    \end{minipage}%
    \begin{minipage}[c]{0.48\columnwidth}
    \includegraphics[scale = 0.36]{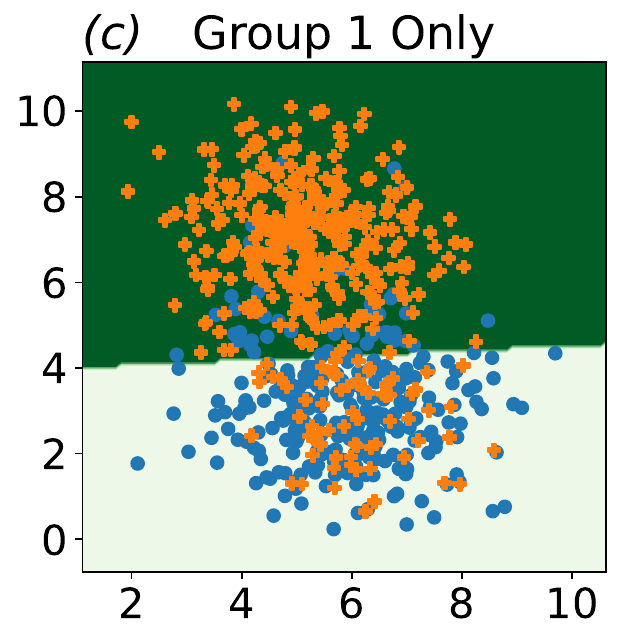}
    \end{minipage}%
    \caption{Dataset generated with group and class-conditional noise and differential subgroup validity. Positive instances ($y^* = 1$) are represented by orange-thickened plus signs, and negative instances ($y^* = 0$) are denoted by blue-filled circles. Figure (a) encompasses all data points, while (b) only includes instances belonging to $g_0$, and (c) only for $g_1$ instances. Group $g_1$, which constitutes 70\% of the whole population, is the majority group. Observed labels for $g_0$ suffer from a high false negative rate, while those for $g_1$ have a high rate of false positives.}
\label{fig: simulation_distribution}
\end{figure}
This simulation, therefore, consists of four clusters depicted in Figure~\ref{fig: simulation_distribution} (b) and (c), where positive observed labels are represented by orange thickened plus signs and negative observed labels are denoted by blue filled circles. (For visualization purposes, we plot a thinner version of the distributions.) Figure~\ref{fig: simulation_distribution} (b) corresponds to group $g_0$ instances, while Figure~\ref{fig: simulation_distribution} (c) corresponds to group $g_1$ instances. Combining Figure~\ref{fig: simulation_distribution} (b) and (c) together, we get the full picture of the four clusters in Figure~\ref{fig: simulation_distribution} (a). 

To elucidate the phenomenon of differential subgroup validity and the rationale behind the decoupling component in DeCoLe, we visualize the linear classifiers that would be learned if the data is pooled together vs. if separate linear classifiers are learned for each group. It is easy to see that the classifiers for group $g_0$ only (Figure~\ref{fig: simulation_distribution} (b)) and for group $g_1$ only (Figure~\ref{fig: simulation_distribution} (c)) exhibit fundamental dissimilarity, being nearly orthogonal. Therefore, when we fit one classifier for both groups, as depicted in Figure~\ref{fig: simulation_distribution} (a), it demonstrates the risks of differential subgroup validity. This phenomenon is closely related to Simpson's Paradox~\citep{simpson1951interpretation}. Note that the classification accuracy for the linear separator in Figure~\ref{fig: simulation_distribution} (a) is notably higher for the majority group compared to the minority group. Furthermore, the classifier in Figure~\ref{fig: simulation_distribution} (a) tends to misclassify positive instances in the minority group $g_0$ as negative and misclassify negative instances in the majority group $g_1$ as positive, reflecting how the classifier may learn and amplify bias in the data labels. If such a classifier is used as a building block of a mislabeling detection algorithm, then such bias will be encoded in it. As we will show in the following subsections, this results in poor mislabeling detection performance when the group-specific nature of predictive relationships and mislabeling patterns is not accounted for. The results also show how this issue is succesfully address by DeCoLe. 
 
\subsubsection{Results and Analysis}\label{Simu_results}
Following the procedure detailed in Algorithm \ref{alg}, we apply DeCoLe to the dataset introduced in Section~\ref{Simu_describ} to perform bias-aware mislabeling detection. We compare the performance of DeCoLe with those of two state-of-the-art alternative frameworks: Confident Learning (CL)~\citep{northcutt2021confident} and Co-Teaching (CoT)~\citep{han2018co}, along with random sampling which provides a useful reference. CL is the most directly related method, as it operates on the same core principles as DeCoLe but assumes noise to be solely class dependent. Co-Teaching, designed for robust learning under high noise, trains two networks in parallel, each selecting small-loss (likely clean) examples to train its peer. To adapt it for mislabeling detection, we use the discarded large-loss instances as its predicted mislabeled set. We use logistic regression as the base model (classifier) for all approaches. We compute 95\% confidence intervals by performing 20 iterations with different random seeds for data shuffling and model initialization to ensure the robustness and consistency of performance evaluation.

\begin{figure}[h]
\centering\includegraphics[scale = 1.9]{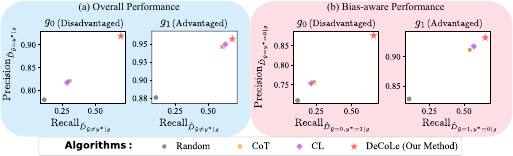} 
\caption{Comparison of DeCoLe and alternative algorithms (Random, CoT, and CL) for bias-aware mislabeling detection, with results disaggregated by demographic groups ($g_0$ and $g_1$).
(a) Overall performance: The x-axis reflects models' ability to recall mislabeled instances (Metric 1), while the y-axis shows the precision of observed labels in data estimated as correctly labeled (Metric 2).
(b) Bias-aware performance: The x-axis reflects the models' ability to recall instances affected by bias-inducing errors (Metric 3), while the y-axis shows the precision of the bias-dominant class in data estimated as correctly labeled (Metric 4). \textit{DeCoLe consistently outperforms competing algorithms according to all metrics.}}
\label{fig:Simu_results}

\end{figure}
We present the results for model performance across different algorithms and demographic groups in Figure~\ref{fig:Simu_results}, using the performance measures defined in Section~\ref{Evaluation}. Specifically, Figure ~\ref{fig:Simu_results} (a) examines the overall mislabeling detection performance, where the x-axis, $\textit{Recall}_{\boldsymbol{\hat {D}}_{\tilde y \neq y^*|g}}$, quantifies each model’s ability to recall mislabeled instances, while the y-axis, $\textit{Precision}_{\boldsymbol{\hat{D}}_{\tilde y = y^*|g}}$, reflects the precision of observed labels in data estimated as correctly labeled. 
Models positioned further to the right demonstrate stronger recall, while higher placement on the y-axis indicates greater precision. DeCoLe consistently occupies the upper-right region of the plots, achieving both the highest recall of mislabeled instances and the highest label precision in data estimated as correclty labeled. 

Figure~\ref{fig:Simu_results} (b) further examines the bias-aware aspect of mislabeling detection by plotting the recall of bias-inducing errors, $\textit{Recall}_{\boldsymbol{\hat {D}}_{\tilde y = c,  y^* = \overline{c}|g}}$, on the x-axis and the precision of the bias-dominant class in the data estimated as correclty labeled, $\textit{Precision}_{\boldsymbol{\hat {D}}_{\tilde y = y^* = c|g}}$ on the y-axis. A position further to the right reflects higher recall of bias-inducing errors, while a higher position indicates greater precision for the bias-dominant class. DeCoLe consistently appears in the upper-right area of the plots, indicating both the highest recall of bias-inducing errors and the highest precision for the bias-dominant class in the data estimated as correctly labeled.

DeCoLe is effective in detecting mislabeled instances and bias-inducing errors to improve data quality, and it yields notably greater performance improvements for group $g_0$---the group that is most adversely affected by label bias in this example---compared to alternative algorithms. Crucially, DeCoLe achieves this without compromising performance for group $g_1$, underscoring DeCoLe’s ability to preserve overall mislabeling detection integrity. This bias-aware advantage is gained through DeCoLe’s ability to account for differential group validity and to perform group-specific mislabeling identification corresponding to the underlying group- and class-conditional noise structure in the dataset.

\begin{table}[H]
    \centering
    \caption{Overall recall of mislabeled instances ($\textit{Recall}_{\boldsymbol{\hat {D}}_{\tilde y \neq y^*|g}}$) using DeCoLe versus competing algorithms. Results are shown for simulations under different error rates, and are disaggregated by demographic groups. Performance values are reported with 95\% confidence intervals (± value). Bold denotes the best-performing model, (*) denotes statistically significant improvements over the second-best algorithm. DeCoLe consistently outperforms competing algorithms.}
    \label{table:Simu_Recall_Overall}
    
    \renewcommand{\arraystretch}{0.8} 
    \setlength{\tabcolsep}{5pt} 
    
    \begin{tabular}{lcccccc}
        \toprule
        $\pi_{0,g_0}$ & \multicolumn{2}{c}{40\%} & \multicolumn{2}{c}{30\%} & \multicolumn{2}{c}{30\%} \\
        \cmidrule(lr){2-3} \cmidrule(lr){4-5} \cmidrule(lr){6-7}
        $\pi_{1,g_1}$ & \multicolumn{2}{c}{10\%} & \multicolumn{2}{c}{10\%} & \multicolumn{2}{c}{20\%} \\
        \midrule
        $g_k$  & $g_0$ & $g_1$ & $g_0$ & $g_1$ & $g_0$ & $g_1$ \\
        \midrule
        \textbf{DeCoLe}  & \textbf{0.817}* \text{\scriptsize$\pm$ 0.011} & \textbf{0.762}* \text{\scriptsize$\pm$ 0.007}  & \textbf{0.696}* \text{\scriptsize$\pm$ 0.015} & \textbf{0.732}* \text{\scriptsize$\pm$ 0.008}& \textbf{0.691}* \text{\scriptsize$\pm$ 0.013} & \textbf{0.703}* \text{\scriptsize$\pm$ 0.010} \\
        CL              & 0.220 \text{\scriptsize$\pm$ 0.007} & 0.719 \text{\scriptsize$\pm$ 0.004} & 0.265 \text{\scriptsize$\pm$ 0.015} & 0.666 \text{\scriptsize$\pm$ 0.014}& 0.336 \text{\scriptsize$\pm$ 0.008} & 0.594 \text{\scriptsize$\pm$ 0.011} \\
        CoT             & 0.226 \text{\scriptsize$\pm$ 0.025} & 0.706 \text{\scriptsize$\pm$ 0.017} & 0.271 \text{\scriptsize$\pm$ 0.031} & 0.644 \text{\scriptsize$\pm$ 0.023} &0.337 \text{\scriptsize$\pm$ 0.028} & 0.567 \text{\scriptsize$\pm$ 0.023}\\
        Random          & 0.139 \text{\scriptsize$\pm$ 0.008} & 0.143 \text{\scriptsize$\pm$ 0.011} & 0.078 \text{\scriptsize$\pm$ 0.007} & 0.077 \text{\scriptsize$\pm$ 0.012} & 0.086 \text{\scriptsize$\pm$ 0.009} & 0.097 \text{\scriptsize$\pm$ 0.010}\\
        \bottomrule
    \end{tabular}
\end{table}

\begin{table}[H]
    \centering
    \caption{Overall precision of the observed label in data estimated as correctly labeled ($\textit{Precision}_{\boldsymbol{\hat{D}}_{\tilde y = y^*|g}}$)  using DeCoLe versus competing algorithms under different error rates. Results are disaggregated by demographic groups. Performance values are reported with 95\% confidence intervals (± value). Bold denotes the best-performing model, while an asterisk (*) denotes statistically significant improvements over the second-best algorithm. DeCoLe consistently outperforms competing algorithms.}
    \label{table:Simu_Acc_Overall}
    
    \renewcommand{\arraystretch}{0.8} 
    \setlength{\tabcolsep}{5 pt} 
    
    \begin{tabular}{lcccccc}
        \toprule
        $\pi_{0,g_0}$ & \multicolumn{2}{c}{40\%} & \multicolumn{2}{c}{30\%} & \multicolumn{2}{c}{30\%} \\
        \cmidrule(lr){2-3} \cmidrule(lr){4-5} \cmidrule(lr){6-7}
        $\pi_{1,g_1}$ & \multicolumn{2}{c}{10\%} & \multicolumn{2}{c}{10\%} & \multicolumn{2}{c}{20\%} \\
        \midrule
        $g_k$  & $g_0$ & $g_1$ & $g_0$ & $g_1$ & $g_0$ & $g_1$ \\
        \midrule
        \textbf{DeCoLe}  & \textbf{0.902}* \text{\scriptsize$\pm$ 0.005} & \textbf{0.947}* \text{\scriptsize$\pm$ 0.002}  & \textbf{0.940}* \text{\scriptsize$\pm$ 0.003} & \textbf{0.978}* \text{\scriptsize$\pm$ 0.001}& \textbf{0.939}* \text{\scriptsize$\pm$ 0.003} & \textbf{0.959}* \text{\scriptsize$\pm$ 0.001} \\
        CL              & 0.664 \text{\scriptsize$\pm$ 0.007} & 0.938 \text{\scriptsize$\pm$ 0.001} & 0.854 \text{\scriptsize$\pm$ 0.003} & 0.973 \text{\scriptsize$\pm$ 0.001} & 0.869 \text{\scriptsize$\pm$ 0.002} & 0.945 \text{\scriptsize$\pm$ 0.001}\\
        CoT             & 0.664 \text{\scriptsize$\pm$ 0.013} & 0.935 \text{\scriptsize$\pm$ 0.004} & 0.854 \text{\scriptsize$\pm$ 0.007} & 0.971 \text{\scriptsize$\pm$ 0.002} & 0.869 \text{\scriptsize$\pm$ 0.006} & 0.941 \text{\scriptsize$\pm$ 0.003} \\
        Random          & 0.626 \text{\scriptsize$\pm$ 0.003} & 0.811 \text{\scriptsize$\pm$ 0.003} & 0.825 \text{\scriptsize$\pm$ 0.004} & 0.923 \text{\scriptsize$\pm$ 0.002} & 0.824 \text{\scriptsize$\pm$ 0.003} & 0.875 \text{\scriptsize$\pm$ 0.002}\\
        \bottomrule
    \end{tabular}
\end{table}


\begin{table}[h]
    \centering
    \caption{Recall of bias-inducing errors ($\textit{Recall}_{\boldsymbol{\hat {D}}_{\tilde y = c,  y^* = \overline{c}|g}}$) using DeCoLe versus competing algorithms under different error rates. Results are disaggregated by demographic groups. Performance values are reported with 95\% confidence intervals (± value). Bold denotes the best-performing model, while an asterisk (*) denotes statistically significant improvements over the second-best algorithm. DeCoLe consistently outperforms competing algorithms.}
    \label{table:Simu_Recall_Bias}
    
    \renewcommand{\arraystretch}{0.8} 
    \setlength{\tabcolsep}{5 pt} 
    
    \begin{tabular}{lcccccc}
        \toprule
        $\pi_{0,g_0}$ & \multicolumn{2}{c}{40\%} & \multicolumn{2}{c}{30\%} & \multicolumn{2}{c}{30\%} \\
        \cmidrule(lr){2-3} \cmidrule(lr){4-5} \cmidrule(lr){6-7}
        $\pi_{1,g_1}$ & \multicolumn{2}{c}{10\%} & \multicolumn{2}{c}{10\%} & \multicolumn{2}{c}{20\%} \\
        \midrule
        $g_k$  & $g_0$ & $g_1$ & $g_0$ & $g_1$ & $g_0$ & $g_1$ \\
        \midrule
        \textbf{DeCoLe}  & \textbf{0.800}* {\scriptsize$\pm$ 0.012} & \textbf{0.725}* {\scriptsize$\pm$ 0.010} & 
                          \textbf{0.664}* {\scriptsize$\pm$ 0.016} & \textbf{0.721}* {\scriptsize$\pm$ 0.010} & 
                          \textbf{0.658}* {\scriptsize$\pm$ 0.013} & \textbf{0.676}* {\scriptsize$\pm$ 0.010} \\
        CL              & 0.146 {\scriptsize$\pm$ 0.008} & 0.664 {\scriptsize$\pm$ 0.009} & 
                          0.167 {\scriptsize$\pm$ 0.012} & 0.593 {\scriptsize$\pm$ 0.017} & 
                          0.244 {\scriptsize$\pm$ 0.009} & 0.529 {\scriptsize$\pm$ 0.011} \\
        CoT             & 0.150 {\scriptsize$\pm$ 0.029} & 0.661 {\scriptsize$\pm$ 0.028} & 
                          0.177 {\scriptsize$\pm$ 0.032} & 0.575 {\scriptsize$\pm$ 0.031} & 
                          0.251 {\scriptsize$\pm$ 0.029} & 0.495 {\scriptsize$\pm$ 0.030} \\
        Random          & 0.139 {\scriptsize$\pm$ 0.008} & 0.140 {\scriptsize$\pm$ 0.013} & 
                          0.079 {\scriptsize$\pm$ 0.008} & 0.076 {\scriptsize$\pm$ 0.014} & 
                          0.087 {\scriptsize$\pm$ 0.009} & 0.100 {\scriptsize$\pm$ 0.011} \\
        \bottomrule
    \end{tabular}
\end{table}


\begin{table}[h]
    \centering
    \caption{Precision of the bias-dominant class $\textit{Precision}_{\boldsymbol{\hat {D}}_{\tilde y = y^* = c|g}}$ across competing algorithms under different error rates. Results are disaggregated by different demographic groups. Performance values are reported with 95\% confidence intervals (± value). Bold denotes the best-performing model, while an asterisk (*) denotes statistically significant improvements over competing methods. DeCoLe consistently outperforms competing algorithms.}
    \label{table:Simu_Acc_Bias}
    
    \renewcommand{\arraystretch}{0.8} 
    \setlength{\tabcolsep}{5 pt} 
    
    \begin{tabular}{lcccccc}
        \toprule
        $\pi_{0,g_0}$ & \multicolumn{2}{c}{40\%} & \multicolumn{2}{c}{30\%} & \multicolumn{2}{c}{30\%} \\
        \cmidrule(lr){2-3} \cmidrule(lr){4-5} \cmidrule(lr){6-7}
        $\pi_{1,g_1}$ & \multicolumn{2}{c}{10\%} & \multicolumn{2}{c}{10\%} & \multicolumn{2}{c}{20\%} \\
        \midrule
        $g_k$  & $g_0$ & $g_1$ & $g_0$ & $g_1$ & $g_0$ & $g_1$ \\
        \midrule
        \textbf{DeCoLe} & \textbf{0.874}* {\scriptsize$\pm$ 0.006} & \textbf{0.921}* {\scriptsize$\pm$ 0.003} & 
                          \textbf{0.905}* {\scriptsize$\pm$ 0.005} & \textbf{0.971}* {\scriptsize$\pm$ 0.002} & 
                          \textbf{0.904}* {\scriptsize$\pm$ 0.005} & \textbf{0.937}* {\scriptsize$\pm$ 0.002} \\
        CL             & 0.618 {\scriptsize$\pm$ 0.007} & 0.906 {\scriptsize$\pm$ 0.002} & 
                          0.792 {\scriptsize$\pm$ 0.005} & 0.958 {\scriptsize$\pm$ 0.003} & 
                          0.809 {\scriptsize$\pm$ 0.003} & 0.910 {\scriptsize$\pm$ 0.001} \\
        CoT            & 0.619 {\scriptsize$\pm$ 0.012} & 0.905 {\scriptsize$\pm$ 0.007} & 
                          0.794 {\scriptsize$\pm$ 0.007} & 0.956 {\scriptsize$\pm$ 0.003} & 
                          0.810 {\scriptsize$\pm$ 0.008} & 0.905 {\scriptsize$\pm$ 0.005} \\
        Random         & 0.580 {\scriptsize$\pm$ 0.006} & 0.763 {\scriptsize$\pm$ 0.005} & 
                          0.761 {\scriptsize$\pm$ 0.005} & 0.902 {\scriptsize$\pm$ 0.003} & 
                          0.760 {\scriptsize$\pm$ 0.006} & 0.828 {\scriptsize$\pm$ 0.004} \\
        \bottomrule
    \end{tabular}
\end{table}

To evaluate the robustness of DeCoLe's performance across diverse settings, we simulate three additional scenarios that vary in group composition and noise structure. In the main setting (Figure~\ref{fig:Simu_results}), we introduced group imbalance: the disadvantaged group $g_0$ constitutes 30\% of the population and is subject to higher label noise ($\pi_{0,{g_0}} = 40\%$ and $\pi_{1,{g_1}} = 20\%$). For the first extended scenario, we assess performance across different error severities. To do this, we vary the values of $\pi_{0,{g_0}}$ and $\pi_{1,{g_1}}$ while maintaining this inequality ($\pi_{0,{g_0}} > \pi_{1,{g_1}}$) and keeping all other settings the same as in the main setting. We report the overall recall of mislabeled instances and the overall precision of observed labels in data estimated as correctly labeled in Tables~\ref{table:Simu_Recall_Overall} and~\ref{table:Simu_Acc_Overall}, respectively. Tables~\ref{table:Simu_Recall_Bias} and~\ref{table:Simu_Acc_Bias} further present the recall of bias-inducing errors and the precision of the bias-dominant class in data estimated as correctly labeled. Across all error rate variations, DeCoLe consistently outperforms competing algorithms in detecting mislabeled and bias-inducing instances while achieving the highest label precision, including for the bias-dominnat class, across both demographic groups. Notably, DeCoLe yields especially strong improvements for $g_0$, the group most affected by label bias, highlighting its effectiveness in bias-aware mislabeling detection.

We further test DeCoLe in two more extended scenarios: First, we relax the assumption that $g_0$ is the minority group by simulating a balanced group setting where $g_0$ and $g_1$ each make up 50\% of the population. We maintain asymmetric error rates such that $\pi_{0,{g_0}} > \pi_{1,{g_1}}$, ensuring that $g_0$ remains the group most affected by label bias, and we vary the severity of the errors. Second, we relax the assumption of asymmetric noise ($\pi_{0,{g_0}} > \pi_{1,{g_1}}$) by simulating a condition in which both groups experience the same error rate ($\pi_{0,{g_0}} = \pi_{1,{g_1}}$), tested at three error rate levels: 20\%, 30\%, and 40\%. For this scenario, we consider $g_0$ to be a minority group that constitutes 30\% of the population. In all cases, DeCoLe maintains superior performance across overall mislabeling detection (Metrics 1 \& 2) and bias-aware performance (Metrics 3 \& 4), reaffirming its robustness and fairness benefits. Detailed results for the extended settings can be found in Appendix~\ref{Appdix:complemt_result1} and~\ref{Appdix:complemt_result2}.

\subsection{Hate Speech Mislabeling Detection} \label{Hate_Speech_describ}
Detecting and mitigating hate speech is an operational challenge for technology platforms~\citep{jahan2023systematic}. Meta, for example, defines hate speech as direct attacks on individuals based on protected characteristics—such as  disability and religion—and enforces policies to remove dehumanizing speech or slurs that target individuals with these characteristics ~\citep{metaHatefulConduct}. Given the vast volume of user-generated content produced online, and the detrimental psychological effects for human moderators who are exposed to disturbing content~\citep{steiger2021}, automated toxicity detection has been recognized as a scalable and cost-effective means to flag possible instances of hate speech ~\citep{jahan2023systematic}. However, the effectiveness of such systems hinges heavily on the quality of the data they rely on throughout the pipeline. When labels indicating whether a post is toxic are biased, analyses and models developed and evaluated on such data are at a higher risk of misclassifying content—either by failing to detect harmful speech or by incorrectly flagging benign posts \citep{sap2019risk, davani2023hate}. Leading platforms such as Meta and YouTube have invested significantly in labeling efforts and automated moderation pipelines. However, a persistent bottleneck remains: the potential bias embedded in the labels themselves~\citep{vidgen2019challenges}. Likewise, labeled hate speech data has become a foundational resource in social science, enabling large-scale quantitative analyses of online behavior \citep{castano2021internet, meza2018targets}. As \cite{castano2021internet} demonstrates, such data are routinely used to support both theoretical modeling and empirical validation in research on online hostility and identity-based violence. However, the presence of biased labels can compromise the integrity of these findings and lead to misguided policy insights. We apply DeCoLe to assess its effectiveness in large-scale bias-aware mislabeling detection as a means of addressing data integrity challenges in hate speech labeling.

To rigorously evaluate mislabeling detection methods like DeCoLe, it is critical to have access to datasets that contain both noisy, real-world labels and higher-quality alternatives—a condition rarely met in practice. Fortunately, such an opportunity arises from the dataset introduced by \cite{kennedy2020constructing}, which provides both commonly used hate speech labels ($\tilde{y}$) and theoretically grounded, more expensive labels ($y^*$), which are estimated on the basis of multiple dimensions of hate speech that are annotated separately, designed specifically to mitigate risks of error and bias in hate speech annotation. 
Specifically, to obtain a gold standard label $y^*$, \cite{kennedy2020constructing} proposes a novel approach based on Rasch Measurement Theory (RMT) to construct a higher quality hate speech label. Their measure articulates hate speech theoretically across eight dimensions (incite violence, humiliate, etc.), capturing the complexity of hate speech and limiting bias stemming from oversimplification and from idiosyncratic definitions of what constitutes hate speech. Furthermore, by evaluating inter-rater reliability, they are able to remove inconsistent raters, correcting human judgment biases and promoting reliability. The researchers assessed the validity and reliability of their proposed measurement approach and found that it demonstrates high internal consistency, test-retest reliability, and construct validity. In summary, \cite{kennedy2020constructing} limits bias in labels for hate speech, but it involves a significantly more costly labeling process, as it requires labels for each instance across eight different dimensions, rather than one. This precludes its widespread use in cost-sensitive settings. The dataset also has the advantage of collecting the more common labels used for hate speech, $\tilde{y}$, by directly asking if a post constitutes hate speech. Thus, the data contains a label $\tilde y$ and an improved label $y^*$ that we leverage to assess the performance of DeCoLe in a real-world setting from an impactful domain.

Beyond the rare opportunity for having this dual label structure, another key strength of this dataset is its rich and fine-grained demographic coverage, with additional annotations indicating demographic information about the target group of the posts, which enables distinct evaluations of label bias detection across a wide range of demographic attributes. The definition of hate speech emphasizes that it involves attacks directed at individuals based on characteristics such as race, gender, or sexuality—but the ways in which such speech manifests can differ significantly across groups. For example, hate speech targeting sexual minorities often relies on coded slurs or moral condemnation, whereas hate speech targeting racial groups may draw from historically rooted stereotypes or dehumanizing language. These distinctions lead to fundamentally different statistical patterns in the data. Thus, each demographic attribute constitutes a different evaluation of the method's performance. Within each broad demographic category (e.g., race), the dataset includes finer-grained groups (e.g., Black, White, Middle Eastern), each exposed to hate speech in contextually unique ways. As a result, the nature and prevalence of label bias along different demographic attributes vary. For example, the manifestation of racial bias in the data is fundamentally different from the manifestation of gender bias. We leverage this rich demographic structure to support multiple evaluations of the proposed methodology. Specifically, we perform separate evaluations, one for each demographic attribute (sexuality, gender, and race), totaling three distinct evaluations.

\subsubsection{Experimental setting} \label{Hate_Speech_Setting} 

\begin{figure}[h]
\vskip -0.02in
\centering\includegraphics[scale = 1.6]{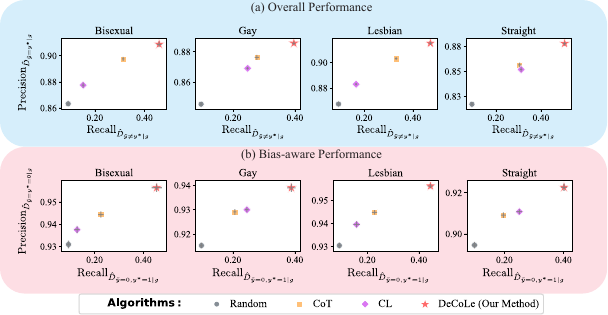} 
\caption{Performance of DeCoLe and alternative algorithms across sexuality groups.
(a) Overall performance: The x-axis reflects models' ability to recall mislabeled instances (Metric 1), while the y-axis shows the precision of observed labels in data estimated as correctly labeled (Metric 2).
(b) Bias-aware performance: The x-axis reflects the models' ability to recall instances affected by bias-inducing errors (Metric 3), while the y-axis shows the precision of the bias-dominant class in data estimated as correctly labeled (Metric 4). Each panel corresponds to a distinct sexuality group. DeCoLe consistently achieves highest recall and precision among all alternatives, demonstrating its effectiveness in bias-aware mislabeling detection.}
\label{fig:MHS_Sexuality_Fig}
\vskip -0.02in
\end{figure}


To empirically evaluate DeCoLe’s ability in bias-aware mislabeling detection in hate speech data, we leverage the rare dual-label structure of the dataset introduced by \cite{kennedy2020constructing}. Specifically, we treat the commonly used hate speech labels derived from a single survey item as observed labels ($\tilde{y}$), and use the theory-grounded, expert-curated annotations as the gold standard labels ($y^*$). 

We include demographic groups with at least 3,000 instances in the evaluation set, to ensure sufficient data for training deep neural networks while using high-dimensional word embeddings, and enable group-specific evaluations with reliable statistical power. We perform three separate evaluations—one each for sexuality, race, and gender. Since hate speech detection is a Natural Language Processing (NLP) task, for every evaluated algorithm, we use Neural Network as the base model and transform each posts into a 768 dimensions feature vector under DistilBERT embeddings, chosen for their ability to capture nuanced semantic relationships while ensuring computational efficiency. To ensure robustness, we performed five runs of data shuffling for 60\%-40\% train-test partitions, each with model initialization with different random seeds per algorithm and report 95\% confidence bounds. %

\subsubsection{Empirical Results} \label{Hate_Speech_Results}
Our empirical findings demonstrate that DeCoLe outperforms competing algorithms by significantly improving the overall recall of the mislabeled instances, the recall of the bias-inducing error, and by more effectively improving the overall precision of observed labels and the precision of the bias-dominant class in the data estimated as correctly labeled. 

\begin{figure}[h]
\vskip -0.02in
\centering\includegraphics[scale = 1.4]{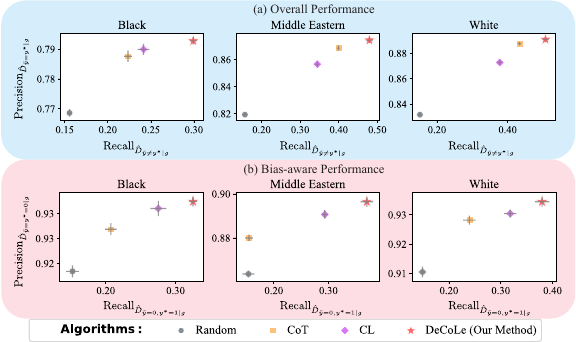} 
\caption{Performance of DeCoLe and competing algorithms in bias-aware mislabeling detection for hate speech annotations on posts targeting different racial groups.
(a) Overall performance: x-axis shows overall recall of mislabeled instances (Metric 1); y-axis shows overall precision of observed labels in data estimated as correctly labeled (Metric 2).
(b) Bias-aware performance: x-axis shows recall of bias-inducing errors (Metric 3); y-axis shows precision of the bias-dominant class in data estimated as correctly labeled (Metric 4).
Each panel corresponds to a distinct racial group. DeCoLe consistently outperforms CoT, CL, and Random Sampling, achieving superior recall and precision across all groups, demonstrating its effectiveness in bias-aware mislabeling detection.}
\label{fig:MHS_Race_Fig}
\vskip -0.01in
\end{figure}

\begin{figure}[h]
\vskip -0.02in
\centering\includegraphics[scale = 1.6]{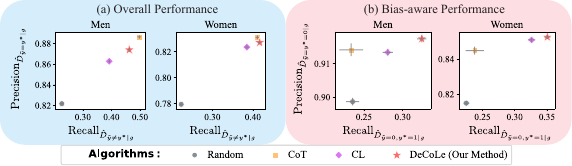} 
\caption{Performance of DeCoLe and competing algorithms in bias-aware mislabeling detection for hate speech annotations on posts targeting different gender groups. 
(a) Overall performance: x-axis shows overall recall of mislabeled instances (Metric 1); y-axis shows overall precision of observed labels in data estimated as correctly labeled (Metric 2).
(b) Bias-aware performance: x-axis shows recall of bias-inducing errors (Metric 3); y-axis shows precision of the bias-dominant class in data estimated as correctly labeled (Metric 4).
Each panel corresponds to a distinct gender group. DeCoLe demonstrates strong performance overall, with particularly notable gains for women. While it ranks second in recall and precision for posts targeting men, it outperforms all alternatives in all other settings.}
\label{fig:MHS_Gender_Fig}
\vskip -0.02in
\end{figure}

Based on the criteria detailed in Section~\ref{Hate_Speech_Setting}, we consider three demographic attributes: sexuality, race, and gender. We separately apply DeCoLe for each of these attributes, yielding three distinct evaluations of the method's performance. Note that each of these identity attributes shapes a distinct relationship between language and labels, with different rate of bias associated to it, enabling evaluation across different manifestations of hate speech. Moreover, this approach enables us to evaluate performance for different values of $K$, i.e., when the number of groups varies. For example, when $g$ refers to sexuality, there are $K=4$ possible values of $g$: $\{\text{bisexual}, \text{gay}, \text{lesbian}, \text{straight}\}$. For race and gender the number of groups that meet the constraint for minimum number of instances specified in Section~\ref{Hate_Speech_Results} are $K=3$ and $K=2$, respectively. We present these three separate sets of results in Figure~\ref{fig:MHS_Sexuality_Fig}, Figure~\ref{fig:MHS_Race_Fig} and Figure ~\ref{fig:MHS_Gender_Fig}, respectively. 

Figure~\ref{fig:MHS_Sexuality_Fig}(a) displays overall performance, with the x-axis presenting the overall recall of mislabeled instances ($\textit{Recall}_{\boldsymbol{\hat {D}}_{\tilde y \neq y^*|g}}$) and the y-axis showing the overall precision of observed labels in data estimated to be correctly labeled ($\textit{Precision}_{\boldsymbol{\hat{D}}_{\tilde y = y^*|g}}$ ). Figure~\ref{fig:MHS_Sexuality_Fig}(b) presents bias-aware performance, plotting the recall of bias-inducing errors ($\textit{Recall}_{\boldsymbol{\hat{D}}{\tilde y = 0, y^* = 1|g}}$) on the x-axis and the precision of observed labels for the bias-dominant class in data estimated as correctly labeled ($\textit{Precision}_{\boldsymbol{\hat{D}}{\tilde y = y^* = 0|g}}$) on the y-axis.  In both subfigures, results are disaggregated by sexuality group, with each panel representing a distinct group. DeCoLe consistently appears in the upper-right portion of the plots across panels, indicating its superior effectiveness in mislabeling detection and its ability to do so in a bias-aware manner.


Moreover, DeCoLe consistently outperforms the competing algorithms in bias-aware mislabeling detection across posts targeting different demographic attributes. Figure~\ref{fig:MHS_Race_Fig} presents the performance of various mislabeling detection algorithms when group membership is defined in terms of racial groups of the targets of posts. Figure~\ref{fig:MHS_Gender_Fig} presents results  when group membership is defined in terms of gender. In both figures, each panel corresponds to a specific group. DeCoLe consistently outperforms competing algorithms in recalling bias-inducing errors and improving the precision of the bias-dominant class in data estimated as correctly labeled across posts targeting both racial and gender groups. In terms of overall performance, DeCoLe remains the top-performing model, surpassing all competing algorithms across settings, except for posts targeting men, where it ranks second in recall of mislabeled instances and precision of observed labels. Overall, the results show that DeCoLe constitutes a state-of-the-art approach for bias-aware mislabeling detection and a valuable tool towards improving data quality and integrity. 


\section{Conclusion and Discussion} \label{conclusion}
Despite growing awareness of the critical role that label quality plays across a wide range of data-driven contexts, and the prevalence of labels that constitute proxies or human judgments prone to labeling bias, 
prior work has not considered methods for bias-aware mislabeling detection, nor empirically evaluated how effectively such mislabels can be detected. We theoretically motivate, develop, and offer in-depth evaluations of Decoupled Confident Learning (DeCoLe)—a principled machine learning framework specifically designed for bias-aware mislabeling detection. Our empirical evaluation in the hate speech domain—a context where label bias has been documented to be pervasive and cause significant harm—demonstrates DeCoLe’s superior performance in bias-aware mislabeling detection when compared to state-of-the-art mislabeling detection algorithms. Given the pressing and persistent data challenges associated with label bias in many important domains, DeCoLe offers a pivotal tool for improving data integrity by effectively performing bias-aware mislabeling detection, facilitating strategic and cost-effective data-auditing and data relabeling efforts, and ultimately enhancing data quality.


\subsection{Managerial Implications}
In an era where data fuels core operations and strategic planning, ensuring data integrity, including label quality, is not just a technical concern but a competitive imperative. Label bias, a widespread form of label flaw, can compromise analytics by skewing insights, leading to poor decisions, lost opportunities, and reputational risks. While most managers recognize the importance of reliable labels in deriving trustworthy insights from AI and analytical systems, there is currently a lack of practical tools to systematically identify mislabeled instances and to support relabeling efforts in a strategic and cost-effective manner. DeCoLe offers a scalable approach to perform bias-aware mislabeling detection and facilitate label quality enhancement before data progresses to versatile operations. By identifying mislabeled instances at the instance level, DeCoLe enables organizations to retain the majority of their existing data while strategically re-labeling or discarding instances that could compromise data quality. This targeted approach supports cost-effective data auditing and improves the overall quality of datasets, and ultimately enhances the reliability of data-driven products. 

\subsubsection{Building on DeCoLe in Data Quality Improvement}
Driven by the need for initiatives that enhance data quality and assist in data preparation, particularly in contexts where label bias may arise, DeCoLe offers a state-of-the-art approach for bias-aware mislabeling detection, supporting such endeavors. DeCoLe allows flexibility in how instances identified as at risk of being labeled are handled. The output of DeCoLe can inform a variety of data quality assessment approaches, such as data cleaning and data relabeling, or more complex algorithmic-driven approaches such as active re-labeling. 

For data cleaning, as illustrated in our experiments, in the presence of labeling bias, pruning instances identified as mislabeled by DeCoLe can effectively reduce the rate of observed mislabeled instances relative to what can be achieved with the state-of-the-art alternatives. In the context of data relabeling, DeCoLe offers a versatile foundation for mitigating label bias across diverse contexts. Its fine-grained identification of mislabeled instances supports integration into human-in-the-loop workflows, enabling experts or labeling panels to focus review efforts on high-risk instances. This is particularly valuable in domains where acquiring high-quality labels is expensive or constrained by other resources such as time, allowing practitioners to allocate annotation resources more strategically. DeCoLe also lends itself to serve as a foundation for algorithmic-driven quality-enhancing approaches. For example, future work can explore its use within active relabeling frameworks that prioritize instances based on their likelihood of being mislabeled and the potential impact of correction on overall fairness~\citep{bernhardt2022active}. These capabilities make DeCoLe well-suited not only for data cleaning and quality assurance but also for guiding strategically allocated expert relabeling efforts. 
By enabling such extensions, DeCoLe contributes not just a standalone solution but a flexible building block that opens a wide array of opportunities for advancing bias-aware data preparation.

\subsection{Limitations and Future Works}
Our empirical evaluation begins with synthetic data and is followed by real-world evidence in the context of hate speech detection. The use of one real-world dataset reflects a broader challenge in the field: the scarcity of datasets that simultaneously provide noisy observed labels, demographic attributes, and higher-quality labels necessary for the assessment of bias-aware mislabeling detection methods. While DeCoLe itself requires only observed labels and demographic attributes for application, effective evaluation demands access to a higher-quality reference—ideally a gold-standard label—to assess the extent of mislabeling and the effectiveness of its detection. The hate speech dataset introduced by \cite{kennedy2020constructing} offers a rare opportunity in this regard. Its theoretically grounded and more costly labeling process yields high-quality reference labels while also including the commonly used, noisier crowdsourced labels.  Moreover, the dataset offers a rare opportunity to conduct evaluations across multiple demographic attributes—such as race, gender, and sexuality—each exhibiting unique linguistic and contextual patterns of hate speech. These differences give rise to distinct forms of label bias, enabling multiple evaluations of the method. Nonetheless, the broader lack of comparable datasets with dual-label structures and rich demographic annotations remains a critical limitation for advancing empirical research in bias-aware mislabeling detection. Future work would benefit from dedicated efforts to construct benchmark datasets for bias-aware mislabeling detection or any efforts facilitating label quality improvement. 

While DeCoLe relaxes the assumption of class-conditioned noise by considering group- and class-conditioned noise structure, it still relies on specific assumptions about the form of label bias present in the data. In particular, we assume that the probability of mislabeling is determined by an instance’s class and group membership.  However, in practice, it is possible that not all instances within the same group and class may have equal likelihoods of being mislabeled—label noise may also be influenced by additional covariates or contextual factors. Our empirical evaluations in hate speech, a domain where label bias is also likely to be dependent on other covariates, show that DeCoLe outperforms alternatives in such a setting. However, future research endeavors could focus on the development of methodologies that provide theoretical guarantees for handling forms of label bias structures beyond group- and class-conditioned noise.

\bibliographystyle{plain}

\bibliography{DeCoLe}

\begin{thebibliography}{10}

\bibitem{abbasi2024pathways}
Ahmed Abbasi, Jeffrey Parsons, Gautam Pant, Olivia R~Liu Sheng, and Suprateek Sarker.
\newblock Pathways for design research on artificial intelligence.
\newblock {\em Inform. Syst. Res.}, 2024.

\bibitem{adams1975management}
Carl~R Adams.
\newblock How management users view information systems.
\newblock {\em Decis. Sci.}, 6(2):337--345, 1975.

\bibitem{agarwal2014big}
Ritu Agarwal and Vasant Dhar.
\newblock Big data, data science, and analytics: The opportunity and challenge for is research.
\newblock {\em Inform. Syst. Res.}, 25(3):443--448, 2014.

\bibitem{akpinar2021effect}
Nil-Jana Akpinar, Maria De-Arteaga, and Alexandra Chouldechova.
\newblock The effect of differential victim crime reporting on predictive policing systems.
\newblock In {\em Proc. 2021 ACM Conf. Fairness, Accountability, and Transparency}, pages 838--849, 2021.

\bibitem{angluin1988learning}
Dana Angluin and Philip Laird.
\newblock Learning from noisy examples.
\newblock {\em Mach. Learn.}, 2:343--370, 1988.

\bibitem{bai2012managing}
Xue Bai, Manuel Nunez, and Jayant~R Kalagnanam.
\newblock Managing data quality risk in accounting information systems.
\newblock {\em Inf. Syst. Res.}, 23(2):453--473, 2012.

\bibitem{ballou1985modeling}
Donald~P Ballou and Harold~L Pazer.
\newblock Modeling data and process quality in multi-input, multi-output information systems.
\newblock {\em Manag. Sci.}, 31(2):150--162, 1985.

\bibitem{barocas2016big}
Solon Barocas and Andrew~D Selbst.
\newblock Big data's disparate impact.
\newblock {\em Calif. L. Rev.}, 104:671, 2016.

\bibitem{bernhardt2022active}
M{\'e}lanie Bernhardt, Daniel~C Castro, Ryutaro Tanno, Anton Schwaighofer, Kerem~C Tezcan, Miguel Monteiro, Shruthi Bannur, Matthew~P Lungren, Aditya Nori, Ben Glocker, et~al.
\newblock Active label cleaning for improved dataset quality under resource constraints.
\newblock {\em Nat. Commun.}, 13(1):1161, 2022.

\bibitem{bjarnadottir2020machine}
Margr{\'e}t~Vilborg Bjarnad{\'o}ttir and David Anderson.
\newblock Machine learning in healthcare: Fairness, issues, and challenges.
\newblock In {\em Pushing the Boundaries: Frontiers in Impactful OR/OM Research}, pages 64--83. INFORMS, 2020.

\bibitem{blodgett2020language}
Su~Lin Blodgett, Solon Barocas, Hal Daum{\'e}~III, and Hanna Wallach.
\newblock Language (technology) is power: A critical survey of" bias" in nlp.
\newblock {\em arXiv preprint arXiv:2005.14050}, 2020.

\bibitem{bolukbasi2016man}
Tolga Bolukbasi, Kai-Wei Chang, James~Y Zou, Venkatesh Saligrama, and Adam~T Kalai.
\newblock Man is to computer programmer as woman is to homemaker? debiasing word embeddings.
\newblock {\em Adv. Neural Inf. Process. Syst.}, 29, 2016.

\bibitem{cai2015challenges}
Li~Cai and Yangyong Zhu.
\newblock The challenges of data quality and data quality assessment in the big data era.
\newblock {\em Data Sci. J.}, 14:2--2, 2015.

\bibitem{cai2022adaptive}
William Cai, Ro~Encarnacion, Bobbie Chern, Sam Corbett-Davies, Miranda Bogen, Stevie Bergman, and Sharad Goel.
\newblock Adaptive sampling strategies to construct equitable training datasets.
\newblock In {\em Proc. 2022 ACM Conf. Fairness, Accountability, and Transparency}, pages 1467--1478, 2022.

\bibitem{caliskan2017semantics}
Aylin Caliskan, Joanna~J Bryson, and Arvind Narayanan.
\newblock Semantics derived automatically from language corpora contain human-like biases.
\newblock {\em Science}, 356(6334):183--186, 2017.

\bibitem{castano2021internet}
Sergio~Andr{\'e}s Casta{\~n}o-Pulgar{\'\i}n, Natalia Su{\'a}rez-Betancur, Luz Magnolia~Tilano Vega, and Harvey Mauricio~Herrera L{\'o}pez.
\newblock Internet, social media and online hate speech. systematic review.
\newblock {\em Aggress. Violent Beh.}, 58:101608, 2021.

\bibitem{chen2019understanding}
Pengfei Chen, Ben~Ben Liao, Guangyong Chen, and Shengyu Zhang.
\newblock Understanding and utilizing deep neural networks trained with noisy labels.
\newblock In {\em International conference on machine learning}, pages 1062--1070. PMLR, 2019.

\bibitem{currim2012modeling}
Faiz Currim and Sudha Ram.
\newblock Modeling spatial and temporal set-based constraints during conceptual database design.
\newblock {\em Inf. Syst. Res.}, 23(1):109--128, 2012.

\bibitem{davani2023hate}
Aida~Mostafazadeh Davani, Mohammad Atari, Brendan Kennedy, and Morteza Dehghani.
\newblock Hate speech classifiers learn normative social stereotypes.
\newblock {\em Trans. Assoc. Comput. Linguist.}, 11:300--319, 2023.

\bibitem{davani2022dealing}
Aida~Mostafazadeh Davani, Mark D{\'\i}az, and Vinodkumar Prabhakaran.
\newblock Dealing with disagreements: Looking beyond the majority vote in subjective annotations.
\newblock {\em Trans. Assoc. Comput. Linguist.}, 10:92--110, 2022.

\bibitem{davenport2006competing}
Thomas~H Davenport et~al.
\newblock Competing on analytics.
\newblock {\em Harv. Bus. Rev.}, 84(1):98, 2006.

\bibitem{dawid1979maximum}
Alexander~Philip Dawid and Allan~M Skene.
\newblock Maximum likelihood estimation of observer error-rates using the em algorithm.
\newblock {\em J. R. Stat. Soc. Ser. C Appl. Stat.}, 28(1):20--28, 1979.

\bibitem{de2022algorithmic}
Maria De-Arteaga, Stefan Feuerriegel, and Maytal Saar-Tsechansky.
\newblock Algorithmic fairness in business analytics: Directions for research and practice.
\newblock {\em Prod. Oper. Manag.}, 31(10):3749--3770, 2022.

\bibitem{dey2010reassessing}
Debabrata Dey and Subodha Kumar.
\newblock Reassessing data quality for information products.
\newblock {\em Manag. Sci.}, 56(12):2316--2322, 2010.

\bibitem{draws2021checklist}
Tim Draws, Alisa Rieger, Oana Inel, Ujwal Gadiraju, and Nava Tintarev.
\newblock A checklist to combat cognitive biases in crowdsourcing.
\newblock In {\em Proc. AAAI Conf. Hum. Comput. Crowdsource}, volume~9, pages 48--59, 2021.

\bibitem{dwork2018decoupled}
Cynthia Dwork, Nicole Immorlica, Adam~Tauman Kalai, and Max Leiserson.
\newblock Decoupled classifiers for group-fair and efficient machine learning.
\newblock In {\em Proc. 2018 ACM Conf. Fairness, Accountability, and Transparency}, pages 119--133. PMLR, 2018.

\bibitem{eickhoff2018cognitive}
Carsten Eickhoff.
\newblock Cognitive biases in crowdsourcing.
\newblock In {\em Proc. 11th ACM Int. Conf. Web Search Data Mining}, pages 162--170, 2018.

\bibitem{elkan2001foundations}
Charles Elkan.
\newblock The foundations of cost-sensitive learning.
\newblock In {\em Int. Jt. Conf. Artif. Intell.}, volume~17, pages 973--978. Lawrence Erlbaum Associates Ltd, 2001.

\bibitem{fogliato2020fairness}
Riccardo Fogliato, Alexandra Chouldechova, and Max G’Sell.
\newblock Fairness evaluation in presence of biased noisy labels.
\newblock In {\em Int. Conf. Artif. Intell. Stat.}, pages 2325--2336. PMLR, 2020.

\bibitem{frenay2013classification}
Beno{\^\i}t Fr{\'e}nay and Michel Verleysen.
\newblock Classification in the presence of label noise: a survey.
\newblock {\em IEEE Trans. Neural Netw. Learn. Syst.}, 25(5):845--869, 2013.

\bibitem{fu2022fair}
Runshan Fu, Manmohan Aseri, Param~Vir Singh, and Kannan Srinivasan.
\newblock “un” fair machine learning algorithms.
\newblock {\em Manage. Sci.}, 68(6):4173--4195, 2022.

\bibitem{fu2020artificial}
Runshan Fu, Yan Huang, and Param~Vir Singh.
\newblock Artificial intelligence and algorithmic bias: Source, detection, mitigation, and implications.
\newblock In {\em Pushing the Boundaries: Frontiers in Impactful OR/OM Research}, pages 39--63. INFORMS, 2020.

\bibitem{ganju2022electronic}
Kartik~K Ganju, Hilal Atasoy, and Paul~A Pavlou.
\newblock Do electronic health record systems increase medicare reimbursements? the moderating effect of the recovery audit program.
\newblock {\em Manage. Sci.}, 68(4):2889--2913, 2022.

\bibitem{goldberger2017training}
Jacob Goldberger and Ehud Ben-Reuven.
\newblock Training deep neural-networks using a noise adaptation layer.
\newblock In {\em International conference on learning representations}, 2017.

\bibitem{gupta2023same}
Soumyajit Gupta, Sooyong Lee, Maria De-Arteaga, and Matthew Lease.
\newblock Same same, but different: Conditional multi-task learning for demographic-specific toxicity detection.
\newblock In {\em Proc. ACM Web Conf. 2023}, pages 3689--3700, 2023.

\bibitem{han2018co}
Bo~Han, Quanming Yao, Xingrui Yu, Gang Niu, Miao Xu, Weihua Hu, Ivor Tsang, and Masashi Sugiyama.
\newblock Co-teaching: Robust training of deep neural networks with extremely noisy labels.
\newblock {\em Adv. Neural Inf. Process. Syst.}, 31, 2018.

\bibitem{hazen2014data}
Benjamin~T Hazen, Christopher~A Boone, Jeremy~D Ezell, and L~Allison Jones-Farmer.
\newblock Data quality for data science, predictive analytics, and big data in supply chain management: An introduction to the problem and suggestions for research and applications.
\newblock {\em Int. J. Prod. Econ.}, 154:72--80, 2014.

\bibitem{hoffman2016racial}
Kelly~M Hoffman, Sophie Trawalter, Jordan~R Axt, and M~Norman Oliver.
\newblock Racial bias in pain assessment and treatment recommendations, and false beliefs about biological differences between blacks and whites.
\newblock {\em Proc. Natl. Acad. Sci.}, 113(16):4296--4301, 2016.

\bibitem{howe2008crowdsourcing}
Jeff Howe.
\newblock {\em Crowdsourcing: How the Power of the Crowd is Driving the Future of Business}.
\newblock Random House, 2008.

\bibitem{hunter1979differential}
John~E Hunter, Frank~L Schmidt, and Ronda Hunter.
\newblock Differential validity of employment tests by race: A comprehensive review and analysis.
\newblock {\em Psychol. Bull.}, 86(4):721, 1979.

\bibitem{jacobs2021measurement}
Abigail~Z Jacobs and Hanna Wallach.
\newblock Measurement and fairness.
\newblock In {\em Proc. 2021 ACM Conf. Fairness, Accountability, and Transparency}, pages 375--385, 2021.

\bibitem{jafar2017emergence}
Musa~J Jafar, Jeffry~Stephen Babb, and Amjad Abdullat.
\newblock Emergence of data analytics in the information systems curriculum.
\newblock {\em Inf. Syst. Educ. J.}, 15(5):22, 2017.

\bibitem{jahan2023systematic}
Md~Saroar Jahan and Mourad Oussalah.
\newblock A systematic review of hate speech automatic detection using natural language processing.
\newblock {\em Neurocomputing}, 546:126232, 2023.

\bibitem{jiang2020identifying}
Heinrich Jiang and Ofir Nachum.
\newblock Identifying and correcting label bias in machine learning.
\newblock In {\em Int. Conf. Artif. Intell. Stat.}, pages 702--712. PMLR, 2020.

\bibitem{jiang2020beyond}
Lu~Jiang, Di~Huang, Mason Liu, and Weilong Yang.
\newblock Beyond synthetic noise: Deep learning on controlled noisy labels.
\newblock In {\em Int. Conf. Mach. Learn.}, pages 4804--4815. PMLR, 2020.

\bibitem{kamiran2009classifying}
Faisal Kamiran and Toon Calders.
\newblock Classifying without discriminating.
\newblock In {\em Proc. 2nd Int. Conf. Comput. Control Commun.}, pages 1--6. IEEE, 2009.

\bibitem{kennedy2020constructing}
Chris~J Kennedy, Geoff Bacon, Alexander Sahn, and Claudia von Vacano.
\newblock Constructing interval variables via faceted rasch measurement and multitask deep learning: a hate speech application.
\newblock {\em arXiv preprint arXiv:2009.10277}, 2020.

\bibitem{kokkodis2021demand}
Marios Kokkodis and Panagiotis~G Ipeirotis.
\newblock Demand-aware career path recommendations: A reinforcement learning approach.
\newblock {\em Manag. Sci.}, 67(7):4362--4383, 2021.

\bibitem{krishnan2005data}
Ramayya Krishnan, James Peters, Rema Padman, and David Kaplan.
\newblock On data reliability assessment in accounting information systems.
\newblock {\em Inf. Syst. Res.}, 16(3):307--326, 2005.

\bibitem{li2022more}
Yunyi Li, Maria De-Arteaga, and Maytal Saar-Tsechansky.
\newblock When more data lead us astray: Active data acquisition in the presence of label bias.
\newblock In {\em Proc. AAAI Conf. Hum. Comput. Crowdsource}, volume~10, pages 133--146, 2022.

\bibitem{li2024label}
Yunyi Li, Maria De-Arteaga, and Maytal Saar-Tsechansky.
\newblock Label bias: A pervasive and invisibilized problem.
\newblock {\em Not. Am. Math. Soc.}, 71(8):1069--1077, 2024.

\bibitem{lipton2018detecting}
Zachary Lipton, Yu-Xiang Wang, and Alexander Smola.
\newblock Detecting and correcting for label shift with black box predictors.
\newblock In {\em International conference on machine learning}, pages 3122--3130. PMLR, 2018.

\bibitem{liu2024financial}
Lisa~Yao Liu.
\newblock Financial statement audits and data breaches.
\newblock {\em Manage. Sci.}, 2024.

\bibitem{lu2020data}
Jing Lu.
\newblock Data analytics research-informed teaching in a digital technologies curriculum.
\newblock {\em INFORMS Transactions on Education}, 20(2):57--72, 2020.

\bibitem{ma2020normalized}
Xingjun Ma, Hanxun Huang, Yisen Wang, Simone Romano, Sarah Erfani, and James Bailey.
\newblock Normalized loss functions for deep learning with noisy labels.
\newblock In {\em Int. Conf. Machine Learning}, pages 6543--6553. PMLR, 2020.

\bibitem{McKinseyDataMatrics}
McKinsey.
\newblock Using customer analytics to boost corporate performance, 2014.

\bibitem{metaHatefulConduct}
{Meta Platforms, Inc.}
\newblock Community standards – hateful conduct.
\newblock \url{https://transparency.meta.com/policies/community-standards/hateful-conduct/}, 2025.
\newblock Accessed: April 30, 2025.

\bibitem{meza2018targets}
Radu~Mihai Meza, Hanna-Orsolya Vincze, and Andreea Mogos.
\newblock Targets of online hate speech in context: a comparative digital social science analysis of comments on public facebook pages from romania and hungary.
\newblock {\em Intersections East Eur. J. Soc. Polit.}, 4(4), 2018.

\bibitem{mitchell2018prediction}
Shira Mitchell, Eric Potash, Solon Barocas, Alexander D'Amour, and Kristian Lum.
\newblock Prediction-based decisions and fairness: A catalogue of choices, assumptions, and definitions.
\newblock {\em arXiv preprint arXiv:1811.07867}, 2018.

\bibitem{mullainathan2021inequity}
Sendhil Mullainathan and Ziad Obermeyer.
\newblock On the inequity of predicting a while hoping for b.
\newblock In {\em AEA Papers and Proceedings}, volume 111, pages 37--42. American Economic Association 2014 Broadway, Suite 305, Nashville, TN 37203, 2021.

\bibitem{natarajan2013learning}
Nagarajan Natarajan, Inderjit~S Dhillon, Pradeep~K Ravikumar, and Ambuj Tewari.
\newblock Learning with noisy labels.
\newblock {\em Adv. Neural Inf. Process. Syst.}, 26, 2013.

\bibitem{nist2023ai_rmf}
NIST.
\newblock Ai risk management framework (ai rmf) playbook, 2023.
\newblock Accessed: 2025-01-08.

\bibitem{northcutt2021confident}
Curtis Northcutt, Lu~Jiang, and Isaac Chuang.
\newblock Confident learning: Estimating uncertainty in dataset labels.
\newblock {\em J. Artif. Intell. Res.}, 70:1373--1411, 2021.

\bibitem{obermeyer2019dissecting}
Ziad Obermeyer, Brian Powers, Christine Vogeli, and Sendhil Mullainathan.
\newblock Dissecting racial bias in an algorithm used to manage the health of populations.
\newblock {\em Science}, 366(6464):447--453, 2019.

\bibitem{otterbacher2015crowdsourcing}
Jahna Otterbacher.
\newblock Crowdsourcing stereotypes: Linguistic bias in metadata generated via gwap.
\newblock In {\em Proc. 33rd Annu. ACM Conf. Hum. Factors Comput. Syst.}, pages 1955--1964, 2015.

\bibitem{parssian2004assessing}
Amir Parssian, Sumit Sarkar, and Varghese~S Jacob.
\newblock Assessing data quality for information products: impact of selection, projection, and cartesian product.
\newblock {\em Manag. Sci.}, 50(7):967--982, 2004.

\bibitem{passi2019problem}
Samir Passi and Solon Barocas.
\newblock Problem formulation and fairness.
\newblock In {\em Proc. 2019 ACM Conf. Fairness, Accountability, and Transparency}, pages 39--48, 2019.

\bibitem{patrini2017making}
Giorgio Patrini, Alessandro Rozza, Aditya Krishna~Menon, Richard Nock, and Lizhen Qu.
\newblock Making deep neural networks robust to label noise: A loss correction approach.
\newblock In {\em Proc. IEEE Conf. Comput. Vision pattern recognition}, pages 1944--1952, 2017.

\bibitem{propublica2016machine}
ProPublica.
\newblock Machine bias, 2016.
\newblock Accessed: 2025-01-28.

\bibitem{sap2019risk}
Maarten Sap, Dallas Card, Saadia Gabriel, Yejin Choi, and Noah~A Smith.
\newblock The risk of racial bias in hate speech detection.
\newblock In {\em Proc. 57th Annu. Meet. Assoc. Comput. Linguist.}, pages 1668--1678, 2019.

\bibitem{shmueli2010explain}
Galit Shmueli.
\newblock To explain or to predict?
\newblock {\em Stat. Sci.}, 25(3):289--310, 2010.

\bibitem{shu2019meta}
Jun Shu, Qi~Xie, Lixuan Yi, Qian Zhao, Sanping Zhou, Zongben Xu, and Deyu Meng.
\newblock Meta-weight-net: Learning an explicit mapping for sample weighting.
\newblock {\em Adv. Neural Inf. Process. Syst.}, 32, 2019.

\bibitem{simpson1951interpretation}
Edward~H Simpson.
\newblock The interpretation of interaction in contingency tables.
\newblock {\em Journal of the Royal Statistical Society: Series B (Methodological)}, 13(2):238--241, 1951.

\bibitem{sjoding2020racial}
Michael~W Sjoding, Robert~P Dickson, Theodore~J Iwashyna, Steven~E Gay, and Thomas~S Valley.
\newblock Racial bias in pulse oximetry measurement.
\newblock {\em N. Engl. J. Med.}, 383(25):2477--2478, 2020.

\bibitem{smyth1994inferring}
Padhraic Smyth, Usama Fayyad, Michael Burl, Pietro Perona, and Pierre Baldi.
\newblock Inferring ground truth from subjective labelling of venus images.
\newblock {\em Adv. Neural Inf. Process. Syst.}, 7, 1994.

\bibitem{snow2008cheap}
Rion Snow, Brendan O’connor, Dan Jurafsky, and Andrew~Y Ng.
\newblock Cheap and fast--but is it good? evaluating non-expert annotations for natural language tasks.
\newblock In {\em Proc. 2008 Conf. Empir. Methods Nat. Lang. Process.}, pages 254--263, 2008.

\bibitem{steiger2021}
Miriah Steiger, Timir~J Bharucha, Sukrit Venkatagiri, Martin~J. Riedl, and Matthew Lease.
\newblock The psychological well-being of content moderators: The emotional labor of commercial moderation and avenues for improving support.
\newblock In {\em Proc. 2021 CHI Conf. Hum. Factors Comput. Syst.}, CHI '21, New York, NY, USA, 2021. Association for Computing Machinery.

\bibitem{sukhbaatar2014training}
Sainbayar Sukhbaatar, Joan Bruna, Manohar Paluri, Lubomir Bourdev, and Rob Fergus.
\newblock Training convolutional networks with noisy labels.
\newblock {\em arXiv preprint arXiv:1406.2080}, 2014.

\bibitem{suresh2021framework}
Harini Suresh and John Guttag.
\newblock A framework for understanding sources of harm throughout the machine learning life cycle.
\newblock In {\em Proc. ACM Conf. Equity Access Algorithms Mechanisms Optimization}, pages 1--9, 2021.

\bibitem{theodorou2025improving}
Brandon Theodorou, Benjamin Danek, Venkat Tummala, Shivam~Pankaj Kumar, Bradley Malin, and Jimeng Sun.
\newblock Improving medical machine learning models with generative balancing for equity and excellence.
\newblock {\em npj Digit. Med.}, 8(1):1--11, 2025.

\bibitem{valiant1984theory}
Leslie~G Valiant.
\newblock A theory of the learnable.
\newblock {\em Commun. ACM}, 27(11):1134--1142, 1984.

\bibitem{van2015learning}
Brendan Van~Rooyen, Aditya Menon, and Robert~C Williamson.
\newblock Learning with symmetric label noise: The importance of being unhinged.
\newblock {\em Advances in neural information processing systems}, 28, 2015.

\bibitem{vidgen2019challenges}
Bertie Vidgen, Alex Harris, Dong Nguyen, Rebekah Tromble, Scott Hale, and Helen Margetts.
\newblock Challenges and frontiers in abusive content detection.
\newblock In {\em Proc. 3rd Workshop Abuse. Lang. Online}. Association for Computational Linguistics, 2019.

\bibitem{wand1996anchoring}
Yair Wand and Richard~Y Wang.
\newblock Anchoring data quality dimensions in ontological foundations.
\newblock {\em Commun. ACM}, 39(11):86--95, 1996.

\bibitem{wang2021fair}
Jialu Wang, Yang Liu, and Caleb Levy.
\newblock Fair classification with group-dependent label noise.
\newblock In {\em Proc. 2021 ACM Conf. Fairness, Accountability, and Transparency}, pages 526--536, 2021.

\bibitem{welinder2010multidimensional}
Peter Welinder, Steve Branson, Pietro Perona, and Serge Belongie.
\newblock The multidimensional wisdom of crowds.
\newblock {\em Adv. Neural Inf. Process. Syst.}, 23, 2010.

\bibitem{whitehill2009whose}
Jacob Whitehill, Ting-fan Wu, Jacob Bergsma, Javier Movellan, and Paul Ruvolo.
\newblock Whose vote should count more: Optimal integration of labels from labelers of unknown expertise.
\newblock {\em Adv. Neural Inf. Process. Syst.}, 22, 2009.

\bibitem{yan2010modeling}
Yan Yan, R{\'o}mer Rosales, Glenn Fung, Mark Schmidt, Gerardo Hermosillo, Luca Bogoni, Linda Moy, and Jennifer Dy.
\newblock Modeling annotator expertise: Learning when everybody knows a bit of something.
\newblock In {\em Proceedings of the thirteenth Int. Conf. Artif. Intell. Stat.}, pages 932--939. JMLR Workshop and Conference Proceedings, 2010.

\bibitem{yu2023unlearning}
Charles Yu, Sullam Jeoung, Anish Kasi, Pengfei Yu, and Heng Ji.
\newblock Unlearning bias in language models by partitioning gradients.
\newblock In {\em Find. Assoc. Comput. Linguist.}, pages 6032--6048, 2023.

\bibitem{zanger2024risk}
Michael Zanger-Tishler, Julian Nyarko, and Sharad Goel.
\newblock Risk scores, label bias, and everything but the kitchen sink.
\newblock {\em Sci. Adv.}, 10(13):eadi8411, 2024.

\bibitem{zelaya2019parametrised}
Vladimiro Zelaya, Paolo Missier, and Dennis Prangle.
\newblock Parametrised data sampling for fairness optimisation.
\newblock {\em KDD XAI}, 2019.

\bibitem{zemel2013learning}
Rich Zemel, Yu~Wu, Kevin Swersky, Toni Pitassi, and Cynthia Dwork.
\newblock Learning fair representations.
\newblock In {\em Int. Conf. Mach. Learn.}, pages 325--333. PMLR, 2013.

\bibitem{zhang2016learning}
Jing Zhang, Xindong Wu, and Victor~S Sheng.
\newblock Learning from crowdsourced labeled data: a survey.
\newblock {\em Artif. Intell. Rev.}, 46:543--576, 2016.

\bibitem{zhao2018learning}
Jieyu Zhao, Yichao Zhou, Zeyu Li, Wei Wang, and Kai-Wei Chang.
\newblock Learning gender-neutral word embeddings.
\newblock {\em arXiv preprint arXiv:1809.01496}, 2018.

\end{thebibliography}

\appendix
\section{Additional Data Generation Details for Controlled Empirical Validation}\label{Appdix:Data_Generation}
We simulate a four-cluster structure based on group and class combinations. The synthetic population contains $N = 10{,}000$ instances with covariates $\boldsymbol{X} \in \mathbb{R}^2$, split into two groups ($g \in {0, 1}$), with group 1 comprising 70\% of the population to reflect group imbalance. Additionally, we account for differential subgroup validity~\citep{hunter1979differential, de2022algorithmic}—a common phenomenon in which the relationship between covariates and target labels varies across groups.  To simulate this, we draw instances for each group–class combination from bivariate normal distributions with group- and class-specific means, while using a shared standard deviation $\sigma = 1.2$ for all distributions. Specifically, for group 0, instances belonging to class 0 ($\boldsymbol{X}_{g=0, y^* = 0}$) are sampled from a bivariate normal distribution $\mathcal{N}((\mu{x_1} = 2, \mu_{x_2} = 3), \sigma^2)$, while class 1 instances ($\boldsymbol{X}_{g=0, y^* = 1}$) are drawn from $\mathcal{N}((\mu{x_1} = 7, \mu_{x_2} = 4), \sigma^2)$. Similarly, for group 1, class 0 instances ($\boldsymbol{X}_{g=1, y^* = 0}$) are drawn from $\mathcal{N}((\mu{x_1} = 6, \mu_{x_2} = 3), \sigma^2)$, and class 1 instances ($\boldsymbol{X}_{g=1, y^* = 1}$) are drawn from $\mathcal{N}((\mu{x_1} = 5, \mu_{x_2} = 7), \sigma^2)$.

\section{Proofs for DeCoLe Theorems} \label{Appdix:Proof}
In this appendix, we present the complete proof supporting our theoretical guarantees.

\textbf{Condition 1: Ideal Predicted Probability}
When the predicted probability $\hat p(\boldsymbol{x}_{g_k})$ is ideal, we have $\hat p(\boldsymbol{x}_{g_k}) = \hat p (\tilde y = 1; \boldsymbol{x}_{g_k}, f_k) = \hat p (\tilde y = 1; \boldsymbol{x}_{g_k} \in \boldsymbol{D}_{y^* = r, g = g_k}, f_k) =  p (\tilde y = 1; \boldsymbol{x}_{g_k} \in \boldsymbol{D}_{y^* = r, g = g_k}, f_k)  =p (\tilde y  = 1| y^*  = r, g = g_k) = p(\boldsymbol{x}_{g_k})$. Essentially, ideal predicted probability $\hat p(\boldsymbol{x}_{g_k})$ match the noise rates $p(\tilde y  = 1| y^*  = r, g = g_k)$ corresponding to true label $r$ for $\boldsymbol{x}_{g_k}$.

\begin{lemma} \label{lemma1}
    For a dataset with biased labels $\boldsymbol{D} \coloneqq (\boldsymbol{x}, \tilde y) ^n$ and group specific classifier $f_k$ for each group $g_k$. Welet $p(\tilde y = 1| y^* = r; g_k)$ denote a shorthand for $ p(\tilde y = 1| y^* = r, g = g_k)$, and adopt this notation analogously for other values and positions of $\tilde y$ and $y^*$, if $\hat p(x_{g_k})$ is ideal, then 
\begin{align}
\text{UB}_{g_k}  =  \sum_{r \in \{0, 1\}}^{} p(\tilde y = 1| y^* = r; g_k) \cdot p(y^* = r | \tilde y = 0; g_k) \\
\text{LB}_{g_k}  =  \sum_{r \in \{0, 1\}}^{} p(\tilde y = 1| y^* = r; g_k) \cdot p(y^* = r | \tilde y = 1; g_k)
\end{align}
\end{lemma}

\textit{Proof.}  We use UB$_{g_k}$ to denote the upper thresholds used to construct the confident negative set for group $g_k$ (CNS$_{g_k}$), and LB$_{g_k}$ to denote the lower thresholds used to construct the confident positive set for group $g_k$ (CPS$_{g_k}$), as defined in Eq.~\ref{eq:ub} and Eq.\ref{eq:lb}. \\ 
By definition,
\begin{align} \label{eq7}
\text{UB}_{g_k}  &= {\mathbb{E}}_{\boldsymbol{x}_{g_k}\in \boldsymbol{D}_{\tilde y = 0}} [\hat p (\tilde y = 1; \boldsymbol{x}_{g_k}, f_k)]
\end{align}
By Bayes Rule, the right-hand side of Eq. \ref{eq7} can be written as: 
\begin{align} \label{eq8}
 \mathop{\mathbb{E}}\limits_{\boldsymbol{x}_{g_k} \in \boldsymbol{X}_{\tilde y = 0 }} \sum_{r \in \{0, 1\}}^{} \hat p (\tilde y = 1|y^* = r; \boldsymbol{x}_{g_k}, f_k) \cdot \hat p (y^* = r; \boldsymbol{x}_{g_k}, f_k)
\end{align}
As the label bias is group and class conditional, Eq. \ref{eq8} can be written as:
\begin{align}
     \sum_{r \in \{0, 1\}}^{} \hat p (\tilde y = 1|y^* = r; g_k) \cdot \mathop{\mathbb{E}}\limits_{\boldsymbol{x}_{g_k} \in \boldsymbol{X}_{\tilde y = 0 }} \hat p (y^* = r; \boldsymbol{x}_{g_k}, f_k)
\end{align}
Applying the ideal predicted probability condition, we get: 
\begin{align}
   \text{UB}_{g_k} &=\sum_{r \in \{0, 1\}}^{} p(\tilde y = 1| y^* = r; g_k) \cdot p(y^* = r | \tilde y = 0; g_k) 
\end{align}

Analogously, applying the same procedure to $\boldsymbol{x}_{g_k} \in \boldsymbol{D}_{\tilde y = 1 }$, we have:
\begin{align}
\text{LB}_{g_k}  =  \sum_{r \in \{0, 1\}}^{} p(\tilde y = 1| y^* = r; g_k) \cdot p(y^* = r | \tilde y = 1; g_k) 
\end{align}

\textbf{Theorem 1}
For a biased dataset $\boldsymbol{D} \coloneqq (\boldsymbol{x}, \tilde y) ^n$, with group- and class-conditional noise $\pi_{0,{g_k}} = P(\tilde y = 0| y^* = 1, g = g_k) < 0.5 $ and $\pi_{1,{g_k}} = P(\tilde y = 1| y^* = 0, g = g_k) <0.5 $, if the classifiers $f_k$ produce ideal predicted probabilities, then the set of detected mislabeled instances $\boldsymbol{\hat {D}}_{\tilde y \neq y^*|g}$ is a consistent estimator of $\boldsymbol{D}_{\tilde y \neq y^*|g}$.

\textit{Proof.} If the set of detected mislabeled instances $\boldsymbol{\hat {D}}_{\tilde y \neq y^*|g}$ is a consistent estimator of $\boldsymbol{D}_{\tilde y \neq y^*|g}$, then, all $ \boldsymbol{x}_{g_k}$ with $\{\tilde y = 0, y^* = 1 \}$ are included in $ \text{CPS}_{g_k} $ and all $ \boldsymbol{x}_{g_k}$ with $ \{\tilde y = 1, y^* = 0 \}$ are included in $\text{CNS}_{g_k} $. Hence, it is sufficient to demonstrate that:
\begin{align}
    \forall \boldsymbol{x}_{g_k} \in \boldsymbol{D}_{\tilde y = 0, y^* = 1}, \hat p(\boldsymbol{x}_{g_k}) \ge  \text{LB}_{g_k}\\
    \forall \boldsymbol{x}_{g_k} \in \boldsymbol{D}_{\tilde y = 1, y^* = 0}, \hat p(\boldsymbol{x}_{g_k}) \le  \text{UB}^*_{g_k}
\end{align}
Given that the predicted probabilities meet the ideal condition, Theorem 1 is substantiated if we can verify the following:
\begin{align}
    \forall \boldsymbol{x}_{g_k} \in \boldsymbol{D}_{\tilde y = 0, y^* = 1}, p(\tilde y  = 1| y^*  = 1, g = g_k) \ge  \text{LB}_{g_k} \label{eq15}\\
    \forall \boldsymbol{x}_{g_k} \in \boldsymbol{D}_{\tilde y = 1, y^* = 0}, p(\tilde y  = 1| y^*  = 0, g = g_k) \le  \text{UB}^*_{g_k} \label{eq14} 
\end{align}
We first prove Eq. \ref{eq15}. Clearly, we have:
\begin{align}  \label{eq16}
    p(\tilde y  = 1| y^*  = 1, g = g_k) \ge p(\tilde y  = 1| y^*  = 1, g = g_k) \cdot 1 
\end{align}
The right-hand side of Eq. \ref{eq16} can be written as:
\begin{align} 
    p(\tilde y  = 1| y^*  = 1, g = g_k) \cdot \sum_{r \in \{0, 1\}}^{}p(y^* = r| \tilde y = 1, g = g_k) \label{eq17}
\end{align}
Rearrange Eq. \ref{eq17} yields:
\begin{align} \label{eq18}
    \sum_{r \in \{0, 1\}}^{} p(\tilde y  = 1| y^*  = 1, g = g_k) \cdot p(y^* = r| \tilde y = 1, g = g_k)
\end{align}
Since $\pi_{1,{g_k} } \le 0.5 $ , we have: $ p(\tilde y  = 1| y^*  = 1, g = g_k) \ge p(\tilde y  = 1| y^*  = 0, g = g_k) $, then, we got that Eq.\ref{eq18} is greater than or equal to the following:
\begin{align} \label{eq19}
     \sum_{r \in \{0, 1\}}^{} p(\tilde y  = 1| y^*  = r, g = g_k) \cdot p(y^* = r| \tilde y = 1, g = g_k)
\end{align}
According to Lemma \ref{lemma1} and based on Eq.\ref{eq16} to Eq.\ref{eq19}, we have:
$p(\tilde y  = 1| y^*  = 1, g = g_k) \ge \text{LB}_{g_k}$.

Thus, we have demonstrated that Eq. \ref{eq15} holds true.  Likewise, following the same procedure $\forall \boldsymbol{x}_{g_k} \in \boldsymbol{X}_{\tilde y = 1, y^* = 0}$, Eq. \ref{eq14} also holds true. As a result, all $ \boldsymbol{x}_{g_k}$ with $\{\tilde y = 0, y^* = 1\}$ are included in $\text{CPS}_{g_k}$ and all $ \boldsymbol{x}_{g_k}$ with $\{\tilde y = 1, y^* = 0\}$ are included in $\text{CNS}_{g_k}$. Therefore, the identified mislabeled set $\boldsymbol{\hat {D}}_{\tilde y \neq y^*|g}$ serves as a consistent estimator of the true mislabeled instances $\boldsymbol{D}_{\tilde y \neq y^*|g}$.

\textbf{Condition 2: Per-example diffracted predicted probability.} The predicted probability $\hat p(\boldsymbol{x}_{g_k})$ provided by model $f_k$ is per-instance diffracted if it follows the relationship $\hat p(\boldsymbol{x}_{g_k}) = p(\boldsymbol{x}_{g_k}) + \epsilon_{\boldsymbol{x}_{g_k}} $ where the noise term  $\epsilon_{\boldsymbol{x}_{g_k}} $ is drawn from the following distribution: $\epsilon_{\boldsymbol{x}_{g_k}} \sim \mathcal{U}[\epsilon_k +\text{LB}^*_{g_k}-p(\boldsymbol{x}_{g_k}), \epsilon_k-\text{LB}^*_{g_k}+ p(\boldsymbol{x}_{g_k})] \hspace{0.2em}$   when $ p(\boldsymbol{x}_{g_k})>\frac{1}{2} $; and  $\epsilon_{\boldsymbol{x}_{g_k}} \sim \mathcal{U}[\epsilon_k -\text{UB}^*_{g_k}+p(\boldsymbol{x}_{g_k}), \epsilon_k+\text{UB}^*_{g_k}-p(\boldsymbol{x}_{g_k})]$ when $p(\boldsymbol{x}_{g_k})<\frac{1}{2} $. Here, $\mathcal{U}$ denotes a uniform distribution, and $\epsilon_k = \mathbb{E}_{\boldsymbol{x_{g_k}} }[\epsilon_{\boldsymbol{x}_{g_k}}] $, where $\epsilon_{\boldsymbol{x}_{g_k}}$ represents the deviation from the ideal predicted probability. $\text{LB}^*_{g_k}$ and $\text{UB}^*_{g_k}$ denote the value of $\text{LB}_{g_k}$ and $\text{UB}_{g_k}$ under condition 1.

\textbf{Theorem 2}
For a biased dataset $\boldsymbol{D} \coloneqq (\boldsymbol{x}, \tilde y) ^n$, with group- and class-conditional noise $\pi_{0,{g_k}} = p(\tilde y = 0| y^* = 1, g = g_k) < 0.5 $ and $\pi_{1,{g_k}} = p(\tilde y = 1| y^* = 0, g = g_k) <0.5 $, if the classifiers $f_k: x_{g_k} \rightarrow \hat p(x_{g_k})$ yield per-instance diffracted predicted probabilities, then the set of detected mislabeled instances $\boldsymbol{\hat {D}}_{\tilde y \neq y^*|g}$ is a consistent estimator of $\boldsymbol{D}_{\tilde y \neq y^*|g}$.

\textbf{Proof} 
Under per-example diffracted predicted probabilities, the predicted probabilities of $\boldsymbol{x}_{g_k}$ are given by:
\begin{align*}
    \hat p(\boldsymbol{x}_{g_k}) = p(\boldsymbol{x}_{g_k}) + \epsilon_{\boldsymbol{x}_{g_k}} 
\end{align*}
where $p(\boldsymbol{x}_{g_k})$ represents the ideal predicted probability introduced in condition 1, and $\hat p(\boldsymbol{x}_{g_k})$ represents $\hat p(\tilde y = 1; \boldsymbol{x}_{g_k}, f_k)$. The error $\epsilon_{\boldsymbol{x}_{g_k}} $ for each group $g_k$ is distributed uniformly as specified in Condition 2. 

Let $\text{LB}^{\epsilon_k}_{g_k}$ and $\text{UB}^{\epsilon_k}_{g_k}$ denote $\text{LB}_{g_k}$ and $\text{UB}_{g_k}$ under condition 2, we have: 
\begin{align}
\text{LB}_{g_k}^{\epsilon_k} &= \frac{1}{|\boldsymbol{D}_{g = g_k, \tilde y = 1}|} \sum_{\boldsymbol{x}_{g_k} \in \boldsymbol{X}_{\tilde y =  1}}^{} p (\boldsymbol{x}_{g_k}) + \epsilon_{\boldsymbol{x}_{g_k}}\\
&= \mathop{\mathbb{E}}\limits_{\boldsymbol{x}_{g_k} \in \boldsymbol{X}_{\tilde y = 1 }}p (\boldsymbol{x}_{g_k}) + \mathop{\mathbb{E}}\limits_{\boldsymbol{x}_{g_k} \in \boldsymbol{X}_{\tilde y = 1 }} \epsilon_{\boldsymbol{x}_{g_k}}
\end{align}

Since $\epsilon_{\boldsymbol{x}_{g_k}}$ is uniformly distributed over $\boldsymbol{x}_{g_k} \in \boldsymbol{X}$, as $n \rightarrow \infty$, $\mathop{\mathbb{E}}\limits_{\boldsymbol{x}_{g_k} \in \boldsymbol{X}_{\tilde y = 1 }} \epsilon_{\boldsymbol{x}_{g_k}} = \mathop{\mathbb{E}}\limits_{\boldsymbol{x}_{g_k} \in \boldsymbol{X}} \epsilon_{\boldsymbol{x}_{g_k}} = \epsilon_k$. Therefore, we have:
\begin{align}
\text{LB}_{g_k}^{\epsilon_k} &= \text{LB}^*_{g_k} + \epsilon_k
\end{align}

From Theorem 1, we have $p (\boldsymbol{x}_{g_k}) \ge \text{LB}_{g_k}^*$. If we can show that:
\begin{align}
p (\boldsymbol{x}_{g_k}) + \epsilon_{\boldsymbol{x}_{g_k}} \ge \text{LB}^*_{g_k} + \epsilon_k \Leftrightarrow p (\boldsymbol{x}_{g_k}) \ge \text{LB}^*_{g_k} 
\end{align}
it means that the confident positive set $\text{CPS}_{g_k}$ created by $\text{LB}_{g_k}^{\epsilon_k}$ are unaltered as compared to $\text{CPS}_{g_k}$ in theorem 1 and therefore $\boldsymbol{\hat {D}}_{\tilde y \neq y^*|g} \overset{\text{Thm.1}}{=} \boldsymbol{D}_{\tilde y \neq y^*|g}$.

According to condition 2, we have:
\begin{align}
    p(\boldsymbol{x}_{g_k}) \ge \frac{1}{2}  \Rightarrow \epsilon_{\boldsymbol{x}_{g_k}} \ge \epsilon_k+\text{LB}^*_{g_k}- p(\boldsymbol{x}_{g_k}) \label{eq26}
\end{align}
Since $p(\boldsymbol{x}_{g_k})\ge\frac{1}{2}$ if an only if $p(\boldsymbol{x}_{g_k}) \ge \text{LB}^*_{g_k}$, it follows that:
\begin{align}
    p(\boldsymbol{x}_{g_k}) \ge \text{LB}^*_{g_k}  \Rightarrow \epsilon_{\boldsymbol{x}_{g_k}} \ge \epsilon_k+\text{LB}^*_{g_k}- p(\boldsymbol{x}_{g_k})
\end{align}
Restructuring, we have
\begin{align}
    p(\boldsymbol{x}_{g_k}) \ge \text{LB}^*_{g_k}  \Rightarrow p(\boldsymbol{x}_{g_k}) + \epsilon_{\boldsymbol{x}_{g_k}} \ge \epsilon_k + \text{LB}^*_{g_k} \label{eq28}
\end{align}

This completes the first half of the proof.
Now we must show the implication in the other direction. According to condition 2, we have:
\begin{align}
    p(\boldsymbol{x}_{g_k}) < \frac{1}{2}  \Rightarrow \epsilon_{\boldsymbol{x}_{g_k}} < \epsilon_k+\text{UB}^*_{g_k}- p(\boldsymbol{x}_{g_k}) \label{eq29}
\end{align}
Since $p(\boldsymbol{x}_{g_k})<\frac{1}{2}$ is equivalent to $p(\boldsymbol{x}_{g_k}) < \text{LB}^*_{g_k}$, and we know that $\text{UB}^*_{g_k} < \text{LB}^*_{g_k}$, then we have:
\begin{align}
    p(\boldsymbol{x}_{g_k}) < \text{LB}^*_{g_k}  \Rightarrow \epsilon_{\boldsymbol{x}_{g_k}} < \epsilon_k+\text{LB}^*_{g_k}- p(\boldsymbol{x}_{g_k}) 
\end{align}
Re-organizing, we have:
\begin{align}
    p(\boldsymbol{x}_{g_k}) < \text{LB}^*_{g_k}  \Rightarrow p(\boldsymbol{x}_{g_k}) + \epsilon_{\boldsymbol{x}_{g_k}} < \epsilon_k+\text{LB}^*_{g_k} 
\end{align}
Using the contrapositive theory, it follows that:
\begin{align}
    p(\boldsymbol{x}_{g_k}) + \epsilon_{\boldsymbol{x}_{g_k}} \ge \epsilon_k+\text{LB}^*_{g_k} \Rightarrow p(\boldsymbol{x}_{g_k}) \ge \text{LB}^*_{g_k} \label{eq32}
\end{align}

Combining Eq. \ref{eq28} and Eq.\ref{eq32}, we have:
\begin{align*}
p (\boldsymbol{x}_{g_k}) + \epsilon_{\boldsymbol{x}_{g_k}} \ge \text{LB}^*_{g_k} + \epsilon_k \Leftrightarrow p (\boldsymbol{x}_{g_k}) \ge \text{LB}^*_{g_k} 
\end{align*}
Therefore,  $\boldsymbol{\hat {D}}_{\tilde y \neq y^*|g} \overset{\text{Thm.1}}{=} \boldsymbol{D}_{\tilde y \neq y^*|g}$. \\
The final step follows from the fact that we have reduced $\boldsymbol{\hat {D}}_{\tilde y \neq y^*|g}$ to identifying instances under the same condition: $p(\boldsymbol{x}_{g_k}) \ge  \text{LB}^*_{g_k}$ as under ideal predicted proabilities. Accordingly, we ensure the accurate identification of label errors and maintain precise estimation. Although we assume a uniform distribution in condition 2, any bounded symmetric distribution with a mode at $\epsilon_k = \mathbb{E}_{\boldsymbol{x} \in \boldsymbol{X}_{g_k}}[\epsilon_{\boldsymbol{x}_{g_k}}] $ is sufficient.

\section{Additional Evaluation Results}

We present comprehensive empirical results for the controlled setting described in Section~\ref{Simu_describ}, evaluating algorithms' performance under varying group compositions, error types, and error rates.  In Appendix~\ref{Appdix:complemt_result1}, we modify the group composition to a balanced distribution—i.e., each group constitutes 50\% of the population—and report results across different values of $\pi_{0_{g_0}}$ and $\pi_{1_{g_1}}$. Appendix~\ref{Appdix:complemt_result2} explores a setting in which both groups suffer from the same error type (false negatives), and presents results under varying error rates $\pi_{0_{g_0}}$ and $\pi_{0_{g_1}}$.
\begin{table}[h]
    \centering
    \caption{Overall recall of mislabeled instances using DeCoLe versus competing approaches under different error rates, disaggregated by demographic groups. The group distribution is balanced, with each gender representing 50\% of the data points. Performance values are reported with 95\% confidence intervals (± value). Bolded values indicate cases where DeCoLe achieves the highest average recall, while an asterisk (*) denotes statistically significant improvements over the second-best algorithm.}
    \label{table:Balanced_group_overall_recall}
    
    \renewcommand{\arraystretch}{0.66} 
    \setlength{\tabcolsep}{3.2pt} 
    
    \begin{tabular}{lcccccc}
        \toprule
        $\pi_{0,g_0}$ & \multicolumn{2}{c}{40\%} & \multicolumn{2}{c}{30\%} & \multicolumn{2}{c}{30\%} \\
        \cmidrule(lr){2-3} \cmidrule(lr){4-5} \cmidrule(lr){6-7}
        $\pi_{1,g_1}$ & \multicolumn{2}{c}{20\%} & \multicolumn{2}{c}{10\%} & \multicolumn{2}{c}{20\%} \\
        \midrule
        $g_k$  & $g_0$ & $g_1$ & $g_0$ & $g_1$ & $g_0$ & $g_1$ \\
        \midrule
        \textbf{DeCoLe}  & \textbf{0.686}* \text{\scriptsize$\pm$ 0.010} & \textbf{0.712}* \text{\scriptsize$\pm$ 0.009}  & \textbf{0.721}* \text{\scriptsize$\pm$ 0.013} & \textbf{0.722}* \text{\scriptsize$\pm$ 0.012} & \textbf{0.708}* \text{\scriptsize$\pm$ 0.011} & \textbf{0.687}* \text{\scriptsize$\pm$ 0.007} \\
        CL              & 0.388 \text{\scriptsize$\pm$ 0.010} & 0.574 \text{\scriptsize$\pm$ 0.008} & 0.403 \text{\scriptsize$\pm$ 0.005} & 0.606 \text{\scriptsize$\pm$ 0.015} & 0.456 \text{\scriptsize$\pm$ 0.009} & 0.528 \text{\scriptsize$\pm$ 0.009} \\
        CoT             & 0.333 \text{\scriptsize$\pm$ 0.016} & 0.618 \text{\scriptsize$\pm$ 0.020} & 0.362 \text{\scriptsize$\pm$ 0.022} & 0.620 \text{\scriptsize$\pm$ 0.019} & 0.429 \text{\scriptsize$\pm$ 0.024} & 0.535 \text{\scriptsize$\pm$ 0.016}\\
        Random          & 0.109 \text{\scriptsize$\pm$ 0.009} & 0.113 \text{\scriptsize$\pm$ 0.008} & 0.098 \text{\scriptsize$\pm$ 0.010} & 0.099 \text{\scriptsize$\pm$ 0.019} & 0.108 \text{\scriptsize$\pm$ 0.013} & 0.101 \text{\scriptsize$\pm$ 0.013}\\
        \bottomrule
    \end{tabular}
\end{table}

\begin{table}[h]
    \centering
    \caption{Overall precision of the observed label in data estimated as correctly labeled using DeCoLe versus competing approaches under different error rates, disaggregated by demographic groups. The group distribution is balanced, with each gender representing 50\% of the data points. Performance values are reported with 95\% confidence intervals (± value). Bolded values indicate cases where DeCoLe achieves the highest average recall, while an asterisk (*) denotes statistically significant improvements over the second-best algorithm.}
    \label{table:Balanced_group_overall_precision}
    
    \renewcommand{\arraystretch}{0.66} 
    \setlength{\tabcolsep}{3.2 pt} 
    
    \begin{tabular}{lcccccc}
        \toprule
        $\pi_{0,g_0}$ & \multicolumn{2}{c}{40\%} & \multicolumn{2}{c}{30\%} & \multicolumn{2}{c}{30\%} \\
        \cmidrule(lr){2-3} \cmidrule(lr){4-5} \cmidrule(lr){6-7}
        $\pi_{1,g_1}$ & \multicolumn{2}{c}{20\%} & \multicolumn{2}{c}{10\%} & \multicolumn{2}{c}{20\%} \\
        \midrule
        $g_k$  & $g_0$ & $g_1$ & $g_0$ & $g_1$ & $g_0$ & $g_1$ \\
        \midrule
        \textbf{DeCoLe}  & \textbf{0.916}* \text{\scriptsize$\pm$ 0.002} & \textbf{0.961}* \text{\scriptsize$\pm$ 0.001}  & \textbf{0.944}* \text{\scriptsize$\pm$ 0.003} & \textbf{0.978}* \text{\scriptsize$\pm$ 0.001} & \textbf{0.942}* \text{\scriptsize$\pm$ 0.002} & \textbf{0.957}* \text{\scriptsize$\pm$ 0.002}\\
        CL              & 0.842 \text{\scriptsize$\pm$ 0.005} & 0.942 \text{\scriptsize$\pm$ 0.002} & 0.881 \text{\scriptsize$\pm$ 0.005} & 0.968 \text{\scriptsize$\pm$ 0.001} & 0.892 \text{\scriptsize$\pm$ 0.004} & 0.935 \text{\scriptsize$\pm$ 0.002}\\
        CoT             & 0.828 \text{\scriptsize$\pm$ 0.006} & 0.948 \text{\scriptsize$\pm$ 0.003} & 0.873 \text{\scriptsize$\pm$ 0.006} & 0.969 \text{\scriptsize$\pm$ 0.002} & 0.887 \text{\scriptsize$\pm$ 0.007} & 0.936 \text{\scriptsize$\pm$ 0.003} \\
        Random          & 0.774 \text{\scriptsize$\pm$ 0.005} & 0.876 \text{\scriptsize$\pm$ 0.003} & 0.825 \text{\scriptsize$\pm$ 0.005} & 0.925 \text{\scriptsize$\pm$ 0.002} & 0.826 \text{\scriptsize$\pm$ 0.005} & 0.875 \text{\scriptsize$\pm$ 0.004}\\
        \bottomrule
    \end{tabular}
\end{table}

\subsection{Balanced Group Distribution and Diverse Error Type for Different Groups} 
\label{Appdix:complemt_result1}
In table ~\ref{table:Balanced_group_overall_recall}, we present the recall of misalbeled instances $\textit{Recall}_{\boldsymbol{\hat {D}}_{\tilde y \neq y^*|g}}$ for Decoupled Confident Learning (DeCoLe - our method), Confident Learning (CL), Co-Teaching (CoT), and Random Sampling (Random), with results disaggregated by demographic groups. Similarly, Table ~\ref{table:Balanced_group_overall_precision} presents the precision of the observed label in data estimated as correctly labeled across DeCoLe and competing algorithms. Performance values are accompanied by their 95\% confidence intervals (± value). Bolded values highlight cases where DeCoLe achieves the highest average accuracy, while an asterisk (*) indicates statistically significant improvements over competing methods. 

In table~\ref{tab:Balanced_group_bias_recall}, we present the recall of bias-inducing errors for Decoupled Confident Learning (DeCoLe - our method), Confident Learning (CL), Co-Teaching (CoT), and Random Sampling (Random), with results disaggregated by demographic groups. Similarly, Table~\ref{tab:Balanced_group_bias_precision} presents the precision of the bias-dominant class across DeCoLe and competing algorithms, also stratified by demographic groups. Performance values are accompanied by their 95\% confidence intervals (± value). Bolded values highlight cases where DeCoLe achieves the highest average accuracy, while an asterisk (*) indicates statistically significant improvements over competing methods. 

\begin{table}[h]
    \centering
    \caption{Recall of bias-inducing errors across DeCoLe and competing methods under different error rates, results are disaggregated by demographic groups. The group distribution is balanced, with each group representing 50\% of the data points. Performance values are reported with 95\% confidence intervals (± value). Bolded values indicate the highest average recall, and an asterisk (*) denotes statistically significant improvements over the second-best algorithm.}
    \label{tab:Balanced_group_bias_recall}
    
    \renewcommand{\arraystretch}{0.66} 
    \setlength{\tabcolsep}{3.2 pt} 
    
    \begin{tabular}{lcccccc}
        \toprule
        $\pi_{0,g_0}$ & \multicolumn{2}{c}{40\%} & \multicolumn{2}{c}{30\%} & \multicolumn{2}{c}{30\%} \\
        \cmidrule(lr){2-3} \cmidrule(lr){4-5} \cmidrule(lr){6-7}
        $\pi_{1,g_1}$ & \multicolumn{2}{c}{20\%} & \multicolumn{2}{c}{10\%} & \multicolumn{2}{c}{20\%} \\
        \midrule
        $g_k$  & $g_0$ & $g_1$ & $g_0$ & $g_1$ & $g_0$ & $g_1$ \\
        \midrule
        \textbf{DeCoLe}  & \textbf{0.654}* \text{\scriptsize$\pm$ 0.011} & \textbf{0.683}* \text{\scriptsize$\pm$ 0.012} & \textbf{0.692}* \text{\scriptsize$\pm$ 0.014} & \textbf{0.696}* \text{\scriptsize$\pm$ 0.017} & \textbf{0.668}* \text{\scriptsize$\pm$ 0.011} & \textbf{0.670}* \text{\scriptsize$\pm$ 0.010} \\
        CL              & 0.320 \text{\scriptsize$\pm$ 0.011} & 0.510 \text{\scriptsize$\pm$ 0.008} & 0.314 \text{\scriptsize$\pm$ 0.008} & 0.462 \text{\scriptsize$\pm$ 0.020} & 0.378 \text{\scriptsize$\pm$ 0.009} & 0.454 \text{\scriptsize$\pm$ 0.013} \\
        CoT             & 0.258 \text{\scriptsize$\pm$ 0.017} & 0.579 \text{\scriptsize$\pm$ 0.022} & 0.268 \text{\scriptsize$\pm$ 0.026} & 0.506 \text{\scriptsize$\pm$ 0.028} & 0.348 \text{\scriptsize$\pm$ 0.026} & 0.465 \text{\scriptsize$\pm$ 0.023} \\
        Random          & 0.110 \text{\scriptsize$\pm$ 0.010} & 0.115 \text{\scriptsize$\pm$ 0.008} & 0.096 \text{\scriptsize$\pm$ 0.010} & 0.101 \text{\scriptsize$\pm$ 0.017} & 0.110 \text{\scriptsize$\pm$ 0.013} & 0.102 \text{\scriptsize$\pm$ 0.016} \\
        \bottomrule
    \end{tabular}
\end{table}

\begin{table}[h]
    \centering
    \caption{Precision of the bias-dominant class in data estimated as correctly labeled using DeCoLe and competing methods under different error rates, disaggregated by demographic groups. The group distribution is balanced, with each group representing 50\% of the population. Performance values are reported with 95\% confidence intervals (± value). Bolded values indicate the highest average precision, and an asterisk (*) denotes statistically significant improvements over the second-best algorithm.}
    \label{tab:Balanced_group_bias_precision}
    
    \renewcommand{\arraystretch}{0.66}
    \setlength{\tabcolsep}{3.2 pt}
    
    \begin{tabular}{lcccccc}
        \toprule
        $\pi_{0,g_0}$ & \multicolumn{2}{c}{40\%} & \multicolumn{2}{c}{30\%} & \multicolumn{2}{c}{30\%} \\
        \cmidrule(lr){2-3} \cmidrule(lr){4-5} \cmidrule(lr){6-7}
        $\pi_{1,g_1}$ & \multicolumn{2}{c}{20\%} & \multicolumn{2}{c}{10\%} & \multicolumn{2}{c}{20\%} \\
        \midrule
        $g_k$  & $g_0$ & $g_1$ & $g_0$ & $g_1$ & $g_0$ & $g_1$ \\
        \midrule
        \textbf{DeCoLe}  & \textbf{0.872}* \text{\scriptsize$\pm$ 0.004} & \textbf{0.939}* \text{\scriptsize$\pm$ 0.002} & \textbf{0.911}* \text{\scriptsize$\pm$ 0.004} & \textbf{0.969}* \text{\scriptsize$\pm$ 0.002} & \textbf{0.905}* \text{\scriptsize$\pm$ 0.004} & \textbf{0.936}* \text{\scriptsize$\pm$ 0.003} \\
        CL              & 0.776 \text{\scriptsize$\pm$ 0.007} & 0.908 \text{\scriptsize$\pm$ 0.003} & 0.821 \text{\scriptsize$\pm$ 0.006} & 0.947 \text{\scriptsize$\pm$ 0.002} & 0.835 \text{\scriptsize$\pm$ 0.006} & 0.899 \text{\scriptsize$\pm$ 0.004} \\
        CoT             & 0.761 \text{\scriptsize$\pm$ 0.007} & 0.920 \text{\scriptsize$\pm$ 0.004} & 0.811 \text{\scriptsize$\pm$ 0.008} & 0.951 \text{\scriptsize$\pm$ 0.003} & 0.828 \text{\scriptsize$\pm$ 0.009} & 0.900 \text{\scriptsize$\pm$ 0.005} \\
        Random          & 0.703 \text{\scriptsize$\pm$ 0.006} & 0.830 \text{\scriptsize$\pm$ 0.006} & 0.759 \text{\scriptsize$\pm$ 0.007} & 0.907 \text{\scriptsize$\pm$ 0.003} & 0.760 \text{\scriptsize$\pm$ 0.007} & 0.828 \text{\scriptsize$\pm$ 0.006} \\
        \bottomrule
    \end{tabular}
\end{table}



\subsection{When the Error Rate for Majority and Minority Groups Are the Same} \label{Appdix:complemt_result2}
\begin{table}[h]
    \centering
    \caption{Overall recall of mislabeled instances using DeCoLe versus competing approaches under settings where both majority (70\%) and minority (30\%) groups experience the same rate of errors (although different error types), we tested three levels of error rates (20\%, 30\%, and 40\%). The results are disaggregated by demographic groups. Performance values are reported with 95\% confidence intervals (± value). Bolded values indicate cases where DeCoLe achieves the highest average recall, while an asterisk (*) denotes statistically significant improvements over the second-best algorithm.}
    \label{tab:Homogeneous_overall_recall}
    
    \renewcommand{\arraystretch}{0.66} 
    \setlength{\tabcolsep}{3.2pt} 
    
    \begin{tabular}{lcccccc}
        \toprule
        $\pi_{0,g_0}$ & \multicolumn{2}{c}{40\%} & \multicolumn{2}{c}{30\%} & \multicolumn{2}{c}{30\%} \\
        \cmidrule(lr){2-3} \cmidrule(lr){4-5} \cmidrule(lr){6-7}
        $\pi_{1, g_1}$ & \multicolumn{2}{c}{20\%} & \multicolumn{2}{c}{10\%} & \multicolumn{2}{c}{20\%} \\
        \midrule
        $g_k$  & $g_0$ & $g_1$ & $g_0$ & $g_1$ & $g_0$ & $g_1$ \\
        \midrule
        \textbf{DeCoLe}  & \textbf{0.726}* \text{\scriptsize$\pm$ 0.018} & \textbf{0.706}* \text{\scriptsize$\pm$ 0.007}  & \textbf{0.696}* \text{\scriptsize$\pm$ 0.012} & \textbf{0.675}* \text{\scriptsize$\pm$ 0.008} & \textbf{0.691}* \text{\scriptsize$\pm$ 0.010} & \textbf{0.691}* \text{\scriptsize$\pm$ 0.007} \\
        CL              &  0.409 \text{\scriptsize$\pm$ 0.015} & 0.586 \text{\scriptsize$\pm$ 0.010} & 0.393 \text{\scriptsize$\pm$ 0.010} & 0.547 \text{\scriptsize$\pm$ 0.008} & 0.410 \text{\scriptsize$\pm$ 0.014} & 0.513 \text{\scriptsize$\pm$ 0.010} \\
        CoT             & 0.409 \text{\scriptsize$\pm$ 0.024} & 0.554 \text{\scriptsize$\pm$ 0.017} &  0.442 \text{\scriptsize$\pm$ 0.043} & 0.489 \text{\scriptsize$\pm$ 0.037} & 0.487 \text{\scriptsize$\pm$ 0.038} & 0.430 \text{\scriptsize$\pm$ 0.035}\\
        Random          & 0.102 \text{\scriptsize$\pm$ 0.015} & 0.095 \text{\scriptsize$\pm$ 0.008} & 0.114 \text{\scriptsize$\pm$ 0.014} & 0.119 \text{\scriptsize$\pm$ 0.007} & 0.133 \text{\scriptsize$\pm$ 0.015} & 0.130 \text{\scriptsize$\pm$ 0.006}\\
        \bottomrule
    \end{tabular}
\end{table}

\begin{table}[h]
    \centering
    \caption{Overall precision of the observed labels in data estimated as correctly labeled across DeCoLe and competing approaches under a homogeneous error type setting, where both majority (70\%) and minority (30\%) groups experience false negative errors, though at different rates. The results are disaggregated by demographic groups. Performance values are reported with 95\% confidence intervals (± value). Bolded values indicate cases where DeCoLe achieves the highest average recall, while an asterisk (*) denotes statistically significant improvements over the second-best algorithm.}
    \label{tab:Homogeneous_overall_precision}
    
    \renewcommand{\arraystretch}{0.66} 
    \setlength{\tabcolsep}{3.2pt} 
    
    \begin{tabular}{lcccccc}
        \toprule
        $\pi_{0,g_0}$ & \multicolumn{2}{c}{40\%} & \multicolumn{2}{c}{30\%} & \multicolumn{2}{c}{30\%} \\
        \cmidrule(lr){2-3} \cmidrule(lr){4-5} \cmidrule(lr){6-7}
        $\pi_{1, g_1}$ & \multicolumn{2}{c}{20\%} & \multicolumn{2}{c}{10\%} & \multicolumn{2}{c}{20\%} \\
        \midrule
        $g_k$  & $g_0$ & $g_1$ & $g_0$ & $g_1$ & $g_0$ & $g_1$  \\
        \midrule
        \textbf{DeCoLe}  & \textbf{0.963}* \text{\scriptsize$\pm$ 0.003} & \textbf{0.960}* \text{\scriptsize$\pm$ 0.001} & \textbf{0.940}*\text{\scriptsize$\pm$ 0.003} & \textbf{0.936}*\text{\scriptsize$\pm$ 0.002}& \textbf{0.918}*\text{\scriptsize$\pm$ 0.003} & \textbf{0.917}*\text{\scriptsize$\pm$ 0.002}\\
        CL              & 0.918 \text{\scriptsize$\pm$ 0.003} & 0.943 \text{\scriptsize$\pm$ 0.001} & 0.880 \text{\scriptsize$\pm$ 0.002} & 0.911 \text{\scriptsize$\pm$ 0.002} & 0.848 \text{\scriptsize$\pm$ 0.004} & 0.873 \text{\scriptsize$\pm$ 0.004}\\
        CoT             &0.918 \text{\scriptsize$\pm$ 0.004} & 0.939 \text{\scriptsize$\pm$ 0.003} & 0.889 \text{\scriptsize$\pm$ 0.010} & 0.900 \text{\scriptsize$\pm$ 0.008} & 0.867 \text{\scriptsize$\pm$ 0.010} & 0.852 \text{\scriptsize$\pm$ 0.009} \\
        Random          & 0.877 \text{\scriptsize$\pm$ 0.003} & 0.875 \text{\scriptsize$\pm$ 0.002} & 0.825 \text{\scriptsize$\pm$ 0.003} & 0.827 \text{\scriptsize$\pm$ 0.004} & 0.776 \text{\scriptsize$\pm$ 0.003} & 0.774 \text{\scriptsize$\pm$ 0.003}\\
        \bottomrule
    \end{tabular}
\end{table}

In table~\ref{tab:Homogeneous_overall_recall}, we present the recall of mislabeled instances for Decoupled Confident Learning (DeCoLe - our method), Confident Learning (CL), Co-Teaching (CoT), and Random Sampling (Random) under a homogeneous error type setting, where both majority and minority groups experience false negative errors, though at different rates. In this setting, the majority group accounts for 70\% of the instances, while the minority group accounts for the remaining 30\%. The results are disaggregated by demographic groups. Similarly, Table~\ref{tab:Homogeneous_overall_precision} presents the precision of the observed labels in data estimated as correctly labeled for DeCoLe and competing algorithms, also stratified by demographic groups. Performance values are accompanied by their 95\% confidence intervals (± value). Bolded values highlight cases where DeCoLe achieves the highest average accuracy, while an asterisk (*) indicates statistically significant improvements over competing methods.

In table~\ref{tab:Homogeneous_bias_recall}, we present the recall of bias-inducing errors for Decoupled Confident Learning (DeCoLe - our method), Confident Learning (CL), Co-Teaching (CoT), and Random Sampling (Random) under a homogeneous error type setting, where both majority and minority groups experience false negative errors, though at different rates. In this setting, the majority group accounts for 70\% of the instances, while the minority group accounts for the remaining 30\%. The results are disaggregated by demographic groups. Similarly, Table~\ref{tab:Homogeneous_bias_precision} presents the accuracy of the bias-dominant class for DeCoLe and competing algorithms, also stratified by demographic groups within this homogeneous error environment. Performance values are accompanied by their 95\% confidence intervals (± value). Bolded values highlight cases where DeCoLe achieves the highest average accuracy, while an asterisk (*) indicates statistically significant improvements over competing methods. 

\begin{table}[h]
    \centering
    \caption{Recall of bias-inducing errors using DeCoLe and competing methods under homogeneous error types across groups. Both majority and minority groups are affected by false negatives. Performance values are reported with 95\% confidence intervals (± value). Bolded values indicate the highest average recall, and an asterisk (*) denotes statistically significant improvements over the second-best algorithm.}
    \label{tab:Homogeneous_bias_recall}
    
    \renewcommand{\arraystretch}{0.66}
    \setlength{\tabcolsep}{3.2 pt}
    
    \begin{tabular}{lcccccc}
        \toprule
        $\pi_{0,g_0}$ & \multicolumn{2}{c}{40\%} & \multicolumn{2}{c}{30\%} & \multicolumn{2}{c}{30\%} \\
        \cmidrule(lr){2-3} \cmidrule(lr){4-5} \cmidrule(lr){6-7}
        $\pi_{1,g_1}$ & \multicolumn{2}{c}{20\%} & \multicolumn{2}{c}{10\%} & \multicolumn{2}{c}{20\%} \\
        \midrule
        $g_k$  & $g_0$ & $g_1$ & $g_0$ & $g_1$ & $g_0$ & $g_1$  \\
        \midrule
        \textbf{DeCoLe}  & \textbf{0.679}* \text{\small$\pm$ 0.016} & \textbf{0.672}* \text{\small$\pm$ 0.009} & \textbf{0.659}*\text{\small$\pm$ 0.013} & \textbf{0.642}*\text{\small$\pm$ 0.010} & \textbf{0.665}*\text{\small$\pm$ 0.009} & \textbf{0.659}*\text{\small$\pm$ 0.006} \\
        CL              & 0.277 \text{\small$\pm$ 0.011} & 0.519 \text{\small$\pm$ 0.008} & 0.321 \text{\small$\pm$ 0.011} & 0.486 \text{\small$\pm$ 0.008} & 0.354 \text{\small$\pm$ 0.015} & 0.455 \text{\small$\pm$ 0.010} \\
        CoT             & 0.288 \text{\small$\pm$ 0.033} & 0.479 \text{\small$\pm$ 0.024} & 0.385 \text{\small$\pm$ 0.048} & 0.415 \text{\small$\pm$ 0.044} & 0.450 \text{\small$\pm$ 0.043} & 0.361 \text{\small$\pm$ 0.042} \\
        Random          & 0.095 \text{\small$\pm$ 0.010} & 0.097 \text{\small$\pm$ 0.010} & 0.114 \text{\small$\pm$ 0.014} & 0.118 \text{\small$\pm$ 0.008} & 0.131 \text{\small$\pm$ 0.013} & 0.129 \text{\small$\pm$ 0.007} \\
        \bottomrule
    \end{tabular}
\end{table}

\begin{table}[h]
    \centering
    \caption{Precision of the bias-dominant class in data estimated as correctly labeled using DeCoLe and competing methods under homogeneous error types across groups. Both majority and minority groups are affected by false negatives. Performance values are reported with 95\% confidence intervals (± value). Bolded values indicate the highest average precision, and an asterisk (*) denotes statistically significant improvements over the second-best algorithm.}
    \label{tab:Homogeneous_bias_precision}
    
    \renewcommand{\arraystretch}{0.66}
    \setlength{\tabcolsep}{3.2 pt}
    
    \begin{tabular}{lcccccc}
        \toprule
        $\pi_{0,g_0}$ & \multicolumn{2}{c}{40\%} & \multicolumn{2}{c}{30\%} & \multicolumn{2}{c}{30\%} \\
        \cmidrule(lr){2-3} \cmidrule(lr){4-5} \cmidrule(lr){6-7}
        $\pi_{1,g_1}$ & \multicolumn{2}{c}{20\%} & \multicolumn{2}{c}{10\%} & \multicolumn{2}{c}{20\%} \\
        \midrule
        $g_k$  & $g_0$ & $g_1$ & $g_0$ & $g_1$ & $g_0$ & $g_1$  \\
        \midrule
        \textbf{DeCoLe}  & \textbf{0.938}* \text{\small$\pm$ 0.004} & \textbf{0.936}* \text{\small$\pm$ 0.002} & \textbf{0.903}*\text{\small$\pm$ 0.004} & \textbf{0.900}*\text{\small$\pm$ 0.004} & \textbf{0.877}*\text{\small$\pm$ 0.004} & \textbf{0.875}*\text{\small$\pm$ 0.004} \\
        CL              & 0.869 \text{\small$\pm$ 0.006} & 0.909 \text{\small$\pm$ 0.002} &  0.823 \text{\small$\pm$ 0.004} & 0.862 \text{\small$\pm$ 0.004} & 0.787 \text{\small$\pm$ 0.006} & 0.814 \text{\small$\pm$ 0.006} \\
        CoT             & 0.871 \text{\small$\pm$ 0.005} & 0.902 \text{\small$\pm$ 0.004} & 0.837 \text{\small$\pm$ 0.012} & 0.846 \text{\small$\pm$ 0.011}& 0.811 \text{\small$\pm$ 0.014} & 0.789 \text{\small$\pm$ 0.011} \\
        Random          & 0.829 \text{\small$\pm$ 0.006} & 0.826 \text{\small$\pm$ 0.003} & 0.759 \text{\small$\pm$ 0.004} & 0.763 \text{\small$\pm$ 0.005} & 0.704 \text{\small$\pm$ 0.004} & 0.704 \text{\small$\pm$ 0.005} \\
        \bottomrule
    \end{tabular}
\end{table}


\end{document}